%% file: root.tex
\documentclass{ifacconf}

\usepackage[table]{xcolor} 

\usepackage{graphicx}      
\usepackage{subcaption}
\usepackage{natbib}        

\usepackage{units, siunitx}
\usepackage{amsmath} 
\usepackage{amssymb} 
\usepackage{mathptmx}

\usepackage[inline]{enumitem}

\usepackage{makecell}
\usepackage{multirow}

\makeatletter
\newcommand*{\romannumber}[1]{\expandafter\@slowromancap\romannumeral #1@}
\makeatother

\usepackage{datatool}
\usepackage{fp}

\usepackage{expl3}

\ExplSyntaxOn
\cs_set_eq:NN \fpeval \fp_eval:n
\ExplSyntaxOff

\usepackage[ruled, algo2e, noend]{algorithm2e}
\SetKwInOut{Parameter}{Parameter}

\usepackage{tikz}
\usepackage{pgfplots}
\usepgfplotslibrary{patchplots}
\usepgfplotslibrary{colorbrewer}
\usepgfplotslibrary{units}

\pgfplotsset{compat=1.7}

\usepgfplotslibrary{external}
\usepackage[textsize=scriptsize]{todonotes}

\usepackage[utf8]{inputenc} 

\newcommand{\figref}[1]{Fig.\,\ref{#1}}
\newcommand{\secref}[1]{Sect.\,\ref{#1}}
\newcommand{\algref}[1]{Algorithm\,\ref{#1}}
\newcommand{\tabref}[1]{Table\,\ref{#1}}
\newcommand{\appref}[1]{Appendix\,\ref{#1}}

\renewcommand{\P}{\mathcal{P}}
\newcommand{\V}{\mathcal{V}}
\newcommand{\D}{\mathcal{D}}
\newcommand{\A}{\mathcal{A}}
\newcommand{\B}{\mathcal{B}}
\newcommand{\R}{\mathcal{R}}
\renewcommand{\O}{\mathcal{O}}

\newcommand{\RR}{\mathbb{R}}

\usepackage{xspace}
\newcommand{\astar}{\ensuremath{\text{A}^{\!\!*}}\xspace}
\newcommand{\worhp}{WORHP\xspace}
\DeclareMathOperator{\dist}{d}
\DeclareMathOperator{\clearance}{\gamma}
\DeclareMathOperator{\expansion}{\xi}
\DeclareMathOperator{\refinements}{\tau}
\DeclareMathOperator{\viewpoint}{\upsilon}

\renewcommand{\d}[1]{\ensuremath{\operatorname{d}\!{#1}}}

\definecolor{b1}{rgb}{0.0, 0.0, 0.6}
\definecolor{b2}{rgb}{0.6, 0.6, 0.9}
\definecolor{b3}{rgb}{0.5, 0.5, 0.8}
\definecolor{b4}{rgb}{0.2, 0.2, 0.7}
\definecolor{k1}{rgb}{0.1, 0.1, 0.1}
\definecolor{k2}{rgb}{0.25, 0.25, 0.25}
\definecolor{k3}{rgb}{0.5,0.5,0.5}
\definecolor{k4}{rgb}{0.9,0.9,0.9}
\definecolor{r1}{rgb}{0.6, 0.0, 0.0}
\definecolor{r2}{rgb}{0.9, 0.6, 0.6}
\definecolor{r3}{rgb}{0.8, 0.5, 0.5}
\definecolor{r4}{rgb}{0.7, 0.2, 0.2}
\definecolor{o1}{rgb}{0.8,0.5,0}
\definecolor{o2}{rgb}{0.5,0.35,0}
\definecolor{g1}{rgb}{0.2,0.8,0.2}
\definecolor{g2}{rgb}{0.1,0.6,0.1}
\definecolor{g3}{rgb}{0.0,0.6,0.0}
\definecolor{g4}{rgb}{0.2,0.7,0.2}
\definecolor{m1}{rgb}{0.8,0.2,0.8}
\definecolor{c1}{rgb}{0.2,0.8,0.8}

\newcommand{\polycolor}{o1}
\newcommand{\datacolor}{k2}
\newcommand{\voronoicolor}{g2}

\DTLnewdb{problemInfo}
\DTLnewdb{trajectoryData}
\newcommand{\PlotTrajectorySearch}[5]{
	
	\DTLifdbexists{problemInfo}{\DTLdeletedb{problemInfo}}{}%
	\DTLloaddb{problemInfo}{#1#3}
	\DTLgetvalue{\vehicleSize}{problemInfo}{1}{1}%
	\DTLgetvalue{\skipFirst}{problemInfo}{2}{1}%
	\DTLgetvalue{\skipSecondRel}{problemInfo}{3}{1}%
	\DTLadd{\skipSecond}{\skipFirst}{\skipSecondRel}
	
	\DTLifdbexists{trajectoryData}{\DTLdeletedb{trajectoryData}}{}%
	\DTLloaddb{trajectoryData}{#1#2}
	\DTLgetvalue{\xpos}{trajectoryData}{\DTLrowcount{trajectoryData}}{2}
	\DTLgetvalue{\ypos}{trajectoryData}{\DTLrowcount{trajectoryData}}{3}
	\DTLgetvalue{\phi}{trajectoryData}{\DTLrowcount{trajectoryData}}{4}
	
	\addplot3 [
	scatter,
	mark=*,
	mark size=1pt,
	point meta min=#4,
	point meta max=#5,
	point meta={z},
	]  table [x expr=\thisrow{x}, y expr=\thisrow{y}, z expr=\thisrow{v}*3.6, col sep=comma] {#1#2};
	
	\draw[line width = \vehiclelinewidth, color=\vehiclecolor, fill=\vehiclecolor, opacity=0.5] (axis cs:\xpos,\ypos) ellipse [x radius=\vehicleSize,  y radius=\vehicleSize];
	\draw[line width = \vehiclelinewidth, color=\vehiclecolor, fill=\vehiclecolor, opacity=0.5] (axis cs:\xpos+\skipFirst*\fpeval{cos(\phi)},\ypos+\skipFirst*\fpeval{sin(\phi)}) ellipse [x radius=\vehicleSize,  y radius=\vehicleSize];
	\draw[line width = \vehiclelinewidth, color=\vehiclecolor, fill=\vehiclecolor, opacity=0.5] (axis cs:\xpos+\skipSecond*\fpeval{cos(\phi)},\ypos+\skipSecond*\fpeval{sin(\phi)}) ellipse [x radius=\vehicleSize,  y radius=\vehicleSize];
	
}

\newcommand{\PlotTrajectorySearchBranches}[2]{
	
	\newread\file
	\openin\file=#1#2
	\loop\unless\ifeof\file
	\read\file to \fileline 
	\if\relax\fileline\relax\else
	\addplot[unbounded coords=discard, line width = \branchlinewidth, color=\branchcolor] table [x index = 1, y index=2, col sep=comma] {"#1\noblankfileline"};
	\fi
	\repeat
	\closein\file
	
}

\newcommand{\PlotVoronoi}[2]{
	
	\newread\file
	\openin\file=#1#2
	\loop\unless\ifeof\file
	\read\file to \fileline 
	\if\relax\fileline\relax\else
	\addplot[unbounded coords=discard, line width = \voronoilinewidth, color=\voronoicolor] table [x expr = \thisrow{x}, y expr=\thisrow{y}, col sep=comma] {"#1\noblankfileline"};
	\fi
	\repeat
	\closein\file
		
}

\newcommand{\PlotSituation}[7]{

	\addplot[on layer=axis background, opacity=0.9] graphics[xmin=\fpeval{#4-10},xmax=\fpeval{#5+10},ymin=\fpeval{#6-10},ymax=\fpeval{#7+10}] {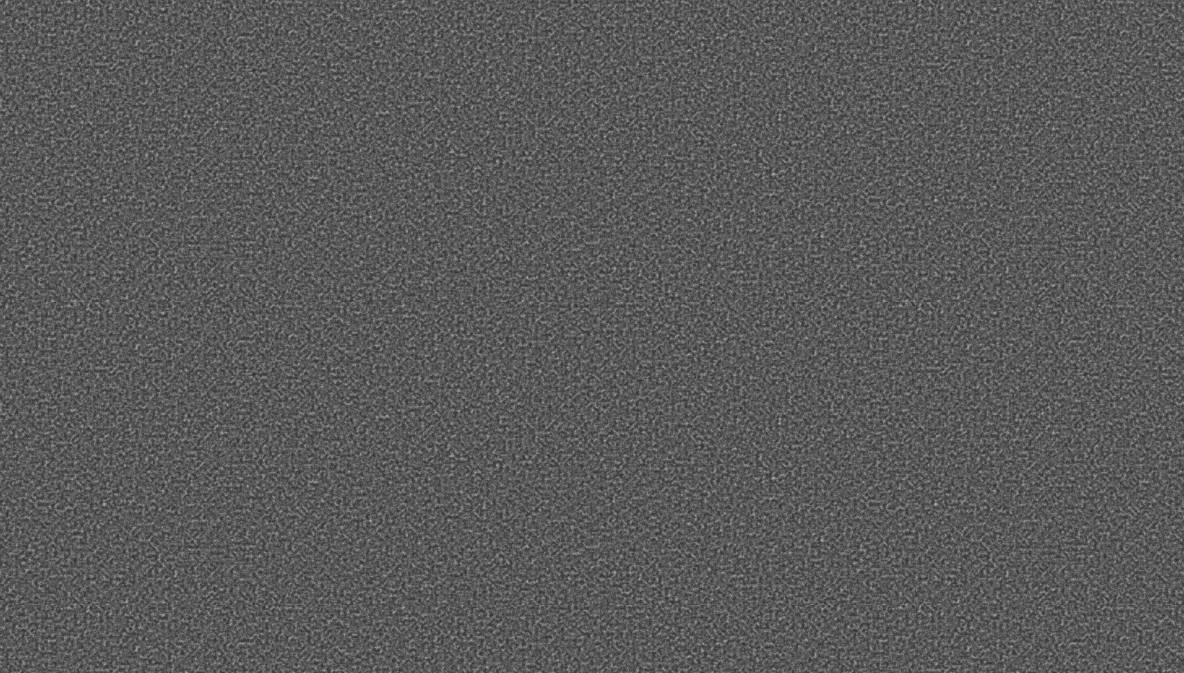};
	
	\addplot[only marks, mark size = \datamarksize, color=\datacolor] table [col sep=comma] {#1#2};
	
	\addplot[unbounded coords=discard, line width = \polylinewidth, color=\polycolor, fill=white, opacity=0.9] table [col sep=comma] {#1#3};
}

\newcommand{\PlotReferenceTraj}[1]{
	\pgfplotstableread{#1}{\data}
	\addplot [b3, very thick, dashed] table [x={x}, y={y}] {\data};
}

\newcommand{\PlotTrajectory}[1]{
	\pgfplotstableread{#1}{\data}
	\addplot [b1, very thick] table [x={x}, y={y}] {\data};
}

\newcommand{\PlotRSTrajectory}[1]{
	\pgfplotstableread{#1}{\data}
	\addplot [c1, very thick, densely dotted] table [x={x}, y={y}] {\data};
}

\DTLnewdb{objectDataZero}
\newcommand{\xspeedZero}{1}
\newcommand{\yspeedZero}{1}
\newcommand{\PlotMovingObjectZero}[2]{
	
	\DTLifdbexists{objectDataZero}{\DTLdeletedb{objectDataZero}}{}%
	
	\DTLloaddb[noheader]{objectDataZero}{#1_0.dat}
	\DTLgetvalue{\objectSizeZero}{objectDataZero}{1}{2}
	\DTLgetvalue{\objectXPosZero}{objectDataZero}{#2}{3}
	\DTLgetvalue{\objectYPosZero}{objectDataZero}{#2}{4}
	
	\DTLgetvalue{\objectZeroTZero}{objectDataZero}{1}{1}
	\DTLgetvalue{\objectZeroTOne}{objectDataZero}{2}{1}
	\DTLgetvalue{\objectZeroXZero}{objectDataZero}{1}{3}
	\DTLgetvalue{\objectZeroXOne}{objectDataZero}{2}{3}
	\DTLgetvalue{\objectZeroYZero}{objectDataZero}{1}{4}
	\DTLgetvalue{\objectZeroYOne}{objectDataZero}{2}{4}	
	
	\renewcommand{\xspeedZero}{\fpeval{(\objectZeroXOne-\objectZeroXZero)/(\objectZeroTOne-\objectZeroTZero)}}
	\renewcommand{\yspeedZero}{\fpeval{(\objectZeroYOne-\objectZeroYZero)/(\objectZeroTOne-\objectZeroTZero)}}
	\draw[line width = \objectlinewidth, color=\speedcolor, fill=\objectcolor] (axis cs:\objectXPosZero,\objectYPosZero) ellipse [x radius=\objectSizeZero,  y radius=\objectSizeZero];		
	
	\draw[->, line width = \speedlinewidth, color=\speedcolor](axis cs:\objectXPosZero,\objectYPosZero)--(axis cs:\objectXPosZero+\xspeedZero*\speedscale,\objectYPosZero+\yspeedZero*\speedscale); 
}

\DTLnewdb{objectDataOne}
\newcommand{\xspeedOne}{1}
\newcommand{\yspeedOne}{1}
\newcommand{\PlotMovingObjectOne}[2]{
	
	\DTLifdbexists{objectDataOne}{\DTLdeletedb{objectDataOne}}{}%
	
	\DTLloaddb[noheader]{objectDataOne}{#1_1.dat}
	\DTLgetvalue{\objectSizeOne}{objectDataOne}{1}{2}
	\DTLgetvalue{\objectXPosOne}{objectDataOne}{#2}{3}
	\DTLgetvalue{\objectYPosOne}{objectDataOne}{#2}{4}
	
	\DTLgetvalue{\objectOneTZero}{objectDataOne}{1}{1}
	\DTLgetvalue{\objectOneTOne}{objectDataOne}{2}{1}
	\DTLgetvalue{\objectOneXZero}{objectDataOne}{1}{3}
	\DTLgetvalue{\objectOneXOne}{objectDataOne}{2}{3}
	\DTLgetvalue{\objectOneYZero}{objectDataOne}{1}{4}
	\DTLgetvalue{\objectOneYOne}{objectDataOne}{2}{4}	
	
	\renewcommand{\xspeedOne}{\fpeval{(\objectOneXOne-\objectOneXZero)/(\objectOneTOne-\objectOneTZero)}}
	\renewcommand{\yspeedOne}{\fpeval{(\objectOneYOne-\objectOneYZero)/(\objectOneTOne-\objectOneTZero)}}

	\draw[line width = \objectlinewidth, color=\speedcolor, fill=\objectcolor] (axis cs:\objectXPosOne,\objectYPosOne) ellipse [x radius=\objectSizeOne,  y radius=\objectSizeOne];		
	
	\draw[->, line width = \speedlinewidth, color=\speedcolor](axis cs:\objectXPosOne,\objectYPosOne)--(axis cs:\objectXPosOne+\xspeedOne*\speedscale,\objectYPosOne+\yspeedOne*\speedscale); 
}

\DTLnewdb{objectDataTwo}
\newcommand{\xspeedTwo}{1}
\newcommand{\yspeedTwo}{1}
\newcommand{\PlotMovingObjectTwo}[2]{
	
	\DTLifdbexists{objectDataTwo}{\DTLdeletedb{objectDataTwo}}{}%
	
	\DTLloaddb[noheader]{objectDataTwo}{#1_2.dat}
	\DTLgetvalue{\objectSizeTwo}{objectDataTwo}{1}{2}
	\DTLgetvalue{\objectXPosTwo}{objectDataTwo}{#2}{3}
	\DTLgetvalue{\objectYPosTwo}{objectDataTwo}{#2}{4}
	
	\DTLgetvalue{\objectTwoTZero}{objectDataTwo}{1}{1}
	\DTLgetvalue{\objectTwoTOne}{objectDataTwo}{2}{1}
	\DTLgetvalue{\objectTwoXZero}{objectDataTwo}{1}{3}
	\DTLgetvalue{\objectTwoXOne}{objectDataTwo}{2}{3}
	\DTLgetvalue{\objectTwoYZero}{objectDataTwo}{1}{4}
	\DTLgetvalue{\objectTwoYOne}{objectDataTwo}{2}{4}	
	
	\renewcommand{\xspeedTwo}{\fpeval{(\objectTwoXOne-\objectTwoXZero)/(\objectTwoTOne-\objectTwoTZero)}}
	\renewcommand{\yspeedTwo}{\fpeval{(\objectTwoYOne-\objectTwoYZero)/(\objectTwoTOne-\objectTwoTZero)}}

	\draw[line width = \objectlinewidth, color=\speedcolor, fill=\objectcolor] (axis cs:\objectXPosTwo,\objectYPosTwo) ellipse [x radius=\objectSizeTwo,  y radius=\objectSizeTwo];		
	
	\draw[->, line width = \speedlinewidth, color=\speedcolor](axis cs:\objectXPosTwo,\objectYPosTwo)--(axis cs:\objectXPosTwo+\xspeedTwo*\speedscale,\objectYPosTwo+\yspeedTwo*\speedscale); 
}

\usepackage[cal=zapfc]{mathalfa}
\newcommand{\C}{\mathcal{C}}

\begin{document}
\begin{frontmatter}
\title{Time-Dependent Hybrid-State \astar and Optimal Control for Autonomous Vehicles in Arbitrary and Dynamic Environments\thanksref{footnoteinfo}}

\thanks[footnoteinfo]{The authors would like to thank the Federal Ministry for Economic Affairs and Energy of Germany (BMWi) and the German Aerospace Center (DLR) Space Administration for supporting this work (grant no. 50\,NA\,1909).}

\author[Uni_HB_O2C]{Andreas Folkers}
\author[Uni_HB_O2C]{Matthias Rick}
\author[Uni_HB_O2C]{Christof Büskens}

\address[Uni_HB_O2C]{Center for Industrial Mathematics, University of Bremen, Germany,\\(e-mail: \{afolkers, mrick, bueskens\}@uni-bremen.de)}

\begin{abstract}                
The development of driving functions for autonomous vehicles in urban environments is still a challenging task. In comparison with driving on motorways, a wide variety of moving road users, such as pedestrians or cyclists, but also the strongly varying and sometimes very narrow road layout pose special challenges. The ability to make fast decisions about exact maneuvers and to execute them by applying sophisticated control commands is one of the key requirements for autonomous vehicles in such situations. In this context we present an algorithmic concept of three correlated methods. Its basis is a novel technique for the automated generation of a free-space polygon, providing a generic representation of the currently drivable area. We then develop a time-dependent hybrid-state \astar algorithm as a model-based planner for the efficient and precise computation of possible driving maneuvers in arbitrary dynamic environments. While on the one hand its results can be used as a basis for making short-term decisions, we also show their applicability as an initial guess for a subsequent trajectory optimization in order to compute applicable control signals. Finally, we provide numerical results for a variety of simulated situations demonstrating the efficiency and robustness of the proposed methods.

\end{abstract}


\end{frontmatter}

\section{Introduction} \label{sec:introduction}
\input{sections/Introduction}

\section{Automated Free-space Polygon Generation} \label{sec:polygon}
\input{sections/Polygon}

\section{Time-dependent hybrid-state $A^*$} \label{sec:hsas}
\input{sections/HSAS}

\section{Optimization \& Optimal Control} \label{sec:optimization}
\input{sections/Optimization}

\section{Numerical Results} \label{sec:results}
\input{sections/Results}

\section{Summary} \label{sec:conclusion}
\input{sections/Conclusion}

\bibliography{references}             

\input{sections/Appendix} \label{app:params}
\end{document}

%% file: sections/Introduction.tex
The vision of driverless cars comes hand in hand with promises of increased road safety or cost efficiency \citep{maurer2016}. While the development of such technologies lasts until today, it already started with early pioneering demonstrations in the 1990s \citep{Dickmanns1997}. However, it is still a challenging task to develop a 
fully autonomous system that is able to safely drive in very general and dynamic environments, such as unrestricted urban areas with other traffic participants. A robust navigation in such situations relies on the one hand on fast planning of possible actions and on the other hand on the computation of corresponding vehicle controls that lead to an efficient and safe but also comfortable maneuver.

Some approaches for the implementation of such strategies introduce highly specialized solutions for specific maneuvers like parking \citep[e.g.][]{parking_dubins_curves}. More general solutions might be expected by using data-driven concepts. Most recently, new methods based on deep learning proposed control policies for tasks like lane following \citep{pilotnet} or even more generic navigation \citep{Folkers2019}. Although these algorithms show promising results, their application in unknown situations strongly depends on the diversity of the data in the training procedure.

A very general approach to solve arbitrary navigation tasks is the usage of optimal control techniques for model-based planning. This method proved to provide very sophisticated solutions in various context such as the control of spacecrafts \citep[e.g.][]{schattel2018diss} or mobile robots \citep[e.g.][]{Meerpohl2019}. Most of all it emerged to be well applicable for the real-time control of autonomous vehicles in the setting of nonlinear model predictive controllers \citep[e.g][]{Falcone2007, Paden2016SurveySelfDriving, Rick2019}.

Certainly, since optimal control methods are usually based on local optimization, their efficient application relies heavily on a well thought out implementation of the corresponding problem. For autonomous driving this is strongly related to the description of spatial constraints given by e.g. static and moving obstacles or the boundary of the lane. In well structured scenarios, which is for example the case in highway driving, methods for obstacle avoidance can be based on geometric considerations \citep[e.g.][]{wolf2008}. However, for more general cases \cite{Meerpohl2019} propose a description of the environment based on free-space polygons. This, once again, has been proven to be well suited for solving optimal control problems in the context of autonomous vehicles \citep{Sommer2018icatt}. Nevertheless, it should be noted that all solutions proposed so far must be provided with a set of viewpoints from which the polygon is then created, which can be a disadvantage in very general applications.

A further crucial element in solving optimal control problems is the requirement of a sufficiently good initial estimate of the solution \citep{Shiller2015}. Ideally, this would take into account both the observed spatial restrictions as well as the constraints on the vehicle's motion, defined by the considered dynamic model. 
The latter condition might be regarded by analytically computing a shortest path according to \cite{reeds1990} as done by \cite{Sommer2018icatt}. However, this does not incorporate any obstacles which might affect the convergence of a local optimizer. 
Alternatively, one could utilize a fast planning algorithm like \astar for finding a reference path in a static environment.
This could be even further enhanced by using, e.g., the hybrid-state \astar method introduced by \cite{dolgov08gppSTAIR}, which also takes a dynamic model of the vehicle into account. It is worth pointing out that, even doing so, this initial guess would still not be able to capture the movement of dynamic traffic participants.

\subsection{Paper Subjects} \label{subsec:contribution}

The contribution of this work is threefold. On the one hand, we will develop an extension of the hybrid-state \astar algorithm to allow fast path planning and decision making in dynamic environments. On the other hand, we show how to use its results as a reference for a consecutive trajectory optimization step which is, in turn, the basis for a model predictive controller. The foundation of both, planning and optimization, lies in the description of the vehicles (arbitrary) static surrounding by a free-space polygon. Therefore, our third contribution is the introduction of a method for its automated generation, which makes this technique applicable as a preliminary step without further knowledge.

\subsection{Research Vehicle} \label{subsec:vehicle}
\begin{figure}[t]
\centering
\includegraphics[width=0.6\linewidth]{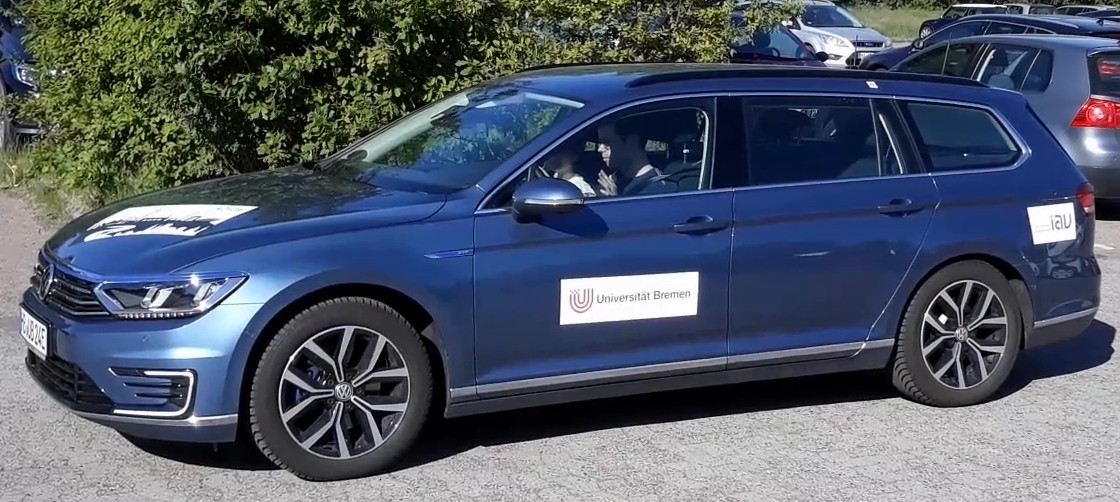}
\caption{The research vehicle.}
\label{fig:vehicle}
\end{figure}

The methods presented in this paper represent a general approach to motion planning and the computation of control commands for self-driving cars in urban environments. However, it should be mentioned that specific vehicle parameters used for the evaluation - like control constraints, signal delays, \mbox{sizes, etc. -} refer to our research vehicle shown in \figref{fig:vehicle}. It allows the execution of autonomous driving maneuvers, for example, by providing corresponding acceleration values and steering angles. See \cite{Rick2019} for more details.

%% file: sections/Polygon.tex
\begin{figure}[t]
	
	\newcommand{\figwidth}{0.27}
	\newcommand{\figheight}{0.15}
	\newcommand{\xmin}{-4}
	\newcommand{\xmax}{25}
	\newcommand{\ymin}{-15}
	\newcommand{\ymax}{25}
	\newcommand{\datamarksize}{0.2}
	\newcommand{\polylinewidth}{1}
	\newcommand{\voronoilinewidth}{0.7}
	
	\pgfplotsset{xtick style={draw=none}, ytick style={draw=none}}
	
	\newcommand\noblankfileline{\expandafter\noblankaux\fileline\relax}
	\def\noblankaux#1 \relax{#1}
	
	\centering
	\begin{tikzpicture}
	\begin{axis}[
	width=\figwidth\textwidth,
	height=\figheight\textheight,
	ylabel= {position [\SI{}{\meter}]},
	xmin = \xmin,
	xmax = \xmax,
	ymin = \ymin,
	ymax = \ymax,
	xtick = \empty,
	]
	
	\addplot[each nth point=2, filter discard warning=false, unbounded coords=discard, only marks, mark size = \datamarksize, color = \datacolor] table [col sep=comma] {data/accessible_area/results_obstacles.dat};
	
	\addplot[unbounded coords=discard, only marks, color=r1] coordinates {(0, 0)};
	
	\end{axis}
	\end{tikzpicture}
	\begin{tikzpicture}
	\begin{axis}[
	width=\figwidth\textwidth,
	height=\figheight\textheight,
	xmin = \xmin,
	xmax = \xmax,
	ymin = \ymin,
	ymax = \ymax,
	xtick = \empty,
	ytick = \empty,
	]
	
	\addplot[each nth point=2, filter discard warning=false, unbounded coords=discard, only marks, mark size = \datamarksize, color=\datacolor] table [col sep=comma] {data/accessible_area/results_obstacles.dat};
	
	\addplot[unbounded coords=discard, line width = \polylinewidth, color=\polycolor] table [col sep=comma] {data/accessible_area/results_poly_0.dat};
	
	\newread\file
	\openin\file=data/accessible_area/results_voronoi_path_0_filelist.dat
		\loop\unless\ifeof\file
			\read\file to \fileline 
			\if\relax\fileline\relax\else
				\addplot[unbounded coords=discard, line width = \voronoilinewidth, color=\voronoicolor] table [x expr = \thisrow{x}, y expr=\thisrow{y}, col sep=comma] {"data/\noblankfileline"};	
			\fi
		\repeat
	\closein\file
	
	\addplot[unbounded coords=discard, only marks, color=r2] coordinates {(0, 0)};
	\addplot[unbounded coords=discard, only marks, color=r1] table [] {data/accessible_area/point_in_poly_peaks0.dat};
	
	\end{axis}
	\end{tikzpicture}
	\\[0.5em]
	\begin{tikzpicture}
	\begin{axis}[
	width=\figwidth\textwidth,
	height=\figheight\textheight,
	ylabel= {position [\SI{}{\meter}]},
	xmin = \xmin,
	xmax = \xmax,
	ymin = \ymin,
	ymax = \ymax,
	]
	
	\addplot[each nth point=2, filter discard warning=false, unbounded coords=discard, only marks, mark size = \datamarksize, color=\datacolor] table [col sep=comma] {data/accessible_area/results_obstacles.dat};	
	
	\addplot[unbounded coords=discard, line width = \polylinewidth, color=\polycolor] table [col sep=comma] {data/accessible_area/results_poly_1.dat};
	
	\newread\file
	\openin\file=data/accessible_area/results_voronoi_path_1_filelist.dat
		\loop\unless\ifeof\file
			\read\file to \fileline 
			\if\relax\fileline\relax\else
				\addplot[unbounded coords=discard, line width = \voronoilinewidth, color=\voronoicolor] table [x expr = \thisrow{x}, y expr=\thisrow{y}, col sep=comma] {"data/\noblankfileline"};	
			\fi
		\repeat
	\closein\file

	\addplot[unbounded coords=discard, only marks, color=r2] coordinates {(0, 0)};
	\addplot[unbounded coords=discard, only marks, color=r2] table [] {data/accessible_area/point_in_poly_peaks0.dat};
	\addplot[unbounded coords=discard, only marks, color=r1] table [] {data/accessible_area/point_in_poly_peaks1.dat};
	
	\end{axis}
	\end{tikzpicture}
	\begin{tikzpicture}
	\begin{axis}[
	width=\figwidth\textwidth,
	height=\figheight\textheight,
	xmin = \xmin,
	xmax = \xmax,
	ymin = \ymin,
	ymax = \ymax,
	ytick = \empty,
	]
	
	\addplot[each nth point=2, filter discard warning=false, unbounded coords=discard, only marks, mark size = \datamarksize, color=\datacolor] table [col sep=comma] {data/accessible_area/results_obstacles.dat};
	
	\addplot[unbounded coords=discard, line width = \polylinewidth, color=\polycolor] table [col sep=comma] {data/accessible_area/results_poly_2.dat};		
	
	\newread\file
	\openin\file=data/accessible_area/results_voronoi_path_2_filelist.dat
		\loop\unless\ifeof\file
			\read\file to \fileline 
			\if\relax\fileline\relax\else
				\addplot[unbounded coords=discard, line width = \voronoilinewidth, color=\voronoicolor] table [x expr = \thisrow{x}, y expr=\thisrow{y}, col sep=comma] {"data/\noblankfileline"};	
			\fi
		\repeat
	\closein\file
	
	\addplot[unbounded coords=discard, only marks, color=r2] coordinates {(0, 0)};
	\addplot[unbounded coords=discard, only marks, color=r2] table [] {data/accessible_area/point_in_poly_peaks0.dat};
	\addplot[unbounded coords=discard, only marks, color=r2] table [] {data/accessible_area/point_in_poly_peaks1.dat};
	\addplot[unbounded coords=discard, only marks, color=r1] table [] {data/accessible_area/point_in_poly_peaks2.dat};
	
	\end{axis}
	\end{tikzpicture}
	\caption{The automated free-space polygon generation algorithm on real data from a parking lot (black). The orange lines show the first three polygons from top left to bottom right. The red dots illustrate the computed viewpoints to be used for the next iteration. Light red dots show previously used viewpoints. The green lines correspond to the respective Voronoi paths $\R$.}
	\label{fig:accessible_area_voronoi}
\end{figure}
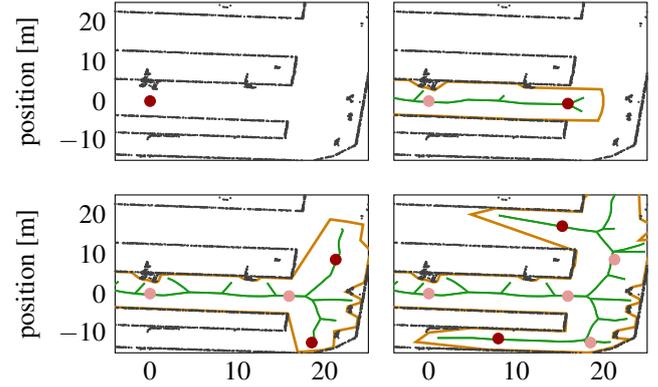

In this section, we describe how a free-space polygon can be generated for the static components of an arbitrary environment.
Even if we take the example of autonomous driving in this work, we would like to emphasize that this method is applicable to general planning problems in the plane.
Thereby, perception might be based on both, measured sensor data as well as a-priori knowledge about fixed parts of the current surrounding (e.g. the position of lanes).
The union of these informations will in the following be denoted as $\D$ and is assumed to be a point cloud to facilitate general applicability.

The algorithm itself can be understood as an iterative process, that increases the size of the free-space polygon in every step.
At each stage the polygon is expanded according to selected viewpoints.
Afterwards new viewpoints are identified based on the latest expansion.
The algorithm is initialized with the systems's current state (position and orientation) $s_0$ being the first viewpoint.

An illustration of the first three steps of an exemplary situation of the whole procedure is given in \figref{fig:accessible_area_voronoi}.
The complete method is presented in \algref{alg:polygon}.

\subsection{Generation of Polygons}
\begin{figure}[b]
	\centering
	\newcommand{\subfigwidth}{0.8\textwidth}
	\begin{subfigure}{0.45\linewidth}
		\centering
		\def\svgwidth{0.9\textwidth}
		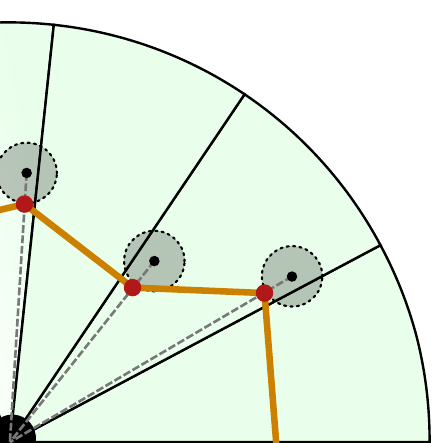
		\caption{Considering the obstacle size.}
		\label{fig:PolyCircleSegmentWithSize}
	\end{subfigure}
	\hfill
	\begin{subfigure}{0.45\linewidth}
		\centering
		\def\svgwidth{0.9\textwidth}
		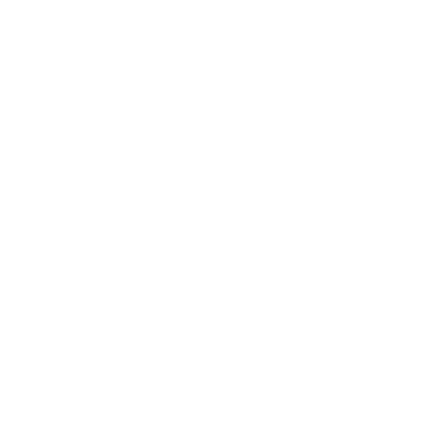
		\caption{Disregarding the obstacle size.}
		\label{fig:PolyCircleSegmentWithoutSize}
	\end{subfigure}
	\caption{Free-Space Polygon Generation. Grey circles represent obstacles points with their specific size and the black dots mark their centers. The dashed lines show the direct connection between obstacle point and viewpoint. Considered obstacles are marked by red dots. Empty circular sectors are colored red, otherwise green. The resulting free-space polygon is given by the orange line.}
\end{figure}
For each given viewpoint $\viewpoint$ a free-space polygon is created and afterwards united with the polygons from previous steps of the automated generation process.
The creation of a single polygon is based on the entries of the obstacle point cloud in the vicinity of the corresponding viewpoint and can be implemented by various methods, e.g. the one intro	duced by \cite{Meerpohl2019}.
Therein, the vertices of a free-space polygon are defined by emitting circles on rays and testing them for collisions with obstacle points.
However, depending on the number of point cloud entries, other approaches may be more efficient.
For example, the amount of necessary evaluations could be reduced by assigning the obstacle points to circular sectors.
In this paper we apply such a \emph{circular sector}-approach which also considers the obstacle point sizes.
For this purpose, the position of the nearest point cloud entry in each sector is reduced to its closest point to the viewpoint, as illustrated in \figref{fig:PolyCircleSegmentWithSize}.
In case of no occupancy, a point is selected according to the maximal predefined expansion $\expansion$.
However, it is possible that an obstacle point extends across more than one sector.
In this case, it is considered in each of them and hence might be marked more than once.
Because of this reason, for instance, obstacle point B is assigned to sectors \romannumber{2} and \romannumber{3} and obstacle point C to sector \romannumber{4} in \figref{fig:PolyCircleSegmentWithSize}.
If obstacle sizes were not taken into account like this, it could happen that a sector would be classified as free, leading to undesired results, as shown in \figref{fig:PolyCircleSegmentWithoutSize}.

\subsection{Generation of Viewpoints}
The selection of valuable new viewpoints has to regard the following two conflicting criteria.
On the one hand, they should be close to the boundary of the current-stage polygon to promote further expansion.
However, on the other hand, if a viewpoint is too close to the boundary, its expansion might be strongly hindered by obstacles directly next to it.

We propose the following procedure to find a good trade-off between both aspects. 
First, local maximizers of the potential function $G_{\P, \V^*}: \RR^2 \rightarrow [-\infty, 0)$ defined by
\begin{equation*}
	G_{\P, \V^*}(p) = 
	\begin{cases}
		- \frac{1}{1+\dist_\P(p)} - \sum\limits_{\viewpoint \in \V^*} \frac{1}{1+\dist(\viewpoint, p)}, &\text{if } p \text{ is inside } \P \\ 
		-\infty, & \text{else}
	\end{cases}
	\label{eq:polygon_potential}
\end{equation*}
are considered as candidates.
Here, $\dist(\cdot, \cdot)$ and $\dist_\P(\cdot)$ define the distances of a point to another one or to the polygon's border, respectively.
$\V^*$ denotes the set of previous viewpoints.
The benefit of this potential is that its maximizers are placed in the center of areas that are far away from old reference points.

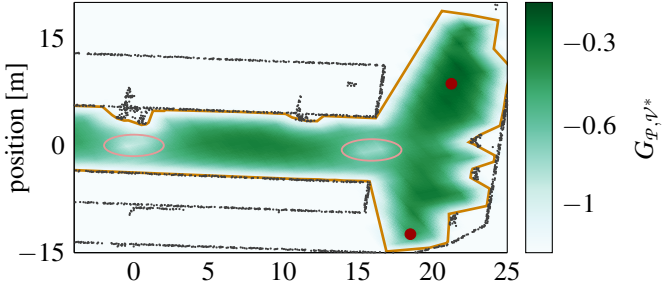
\begin{figure}[t]
	
	\pgfplotsset{xtick style={draw=none}, ytick style={draw=none}}
	
	\newcommand{\datamarksize}{0.2}	
	\newcommand{\xmin}{-4}
	\newcommand{\xmax}{25}
	\newcommand{\ymin}{-15}
	\newcommand{\ymax}{20}
		
	\begin{tikzpicture}
	\begin{axis}[
	width=0.40\textwidth, 
	height=0.2\textheight,	
	xmin = \xmin,
	xmax = \xmax,
	ymin = \ymin,
	ymax = \ymax,
	ylabel= {position [\si{\meter}]},
	ylabel shift = -4mm,
	ytick={15, 0, -15},
	view={0}{90},
	colormap/BuGn,
	colorbar,
	colorbar/width=3.5mm,
	colorbar style={
		ylabel=$G_{\P, \V^*}$,
		xshift=-3.5mm,
		yticklabel=\pgfmathparse{ln(\tick)}\pgfmathprintnumber\pgfmathresult,
		ytick={0.740818221, 0.548811636, 0.367879441}, 
	},
	]
	
	\addplot3[surf,shader=interp, patch type=bilinear] table [x expr=\thisrow{x}, y expr=\thisrow{y}, z expr=exp(\thisrow{z}), col sep=comma] {data/accessible_area/point_in_poly1.dat};
	
	\addplot[each nth point=2, filter discard warning=false, unbounded coords=discard, only marks, mark size = \datamarksize, color=\datacolor] table [col sep=comma] {data/accessible_area/results_obstacles.dat};	
	
	\addplot[unbounded coords=discard, line width = 1, color=o1] table [col sep=comma] {data/accessible_area/results_poly_1.dat};
		
	\addplot[unbounded coords=discard, only marks, color=r1] table [] {data/accessible_area/point_in_poly_peaks1.dat};	
	
	\draw[line width = 0.7, color=r2] (axis cs:0,0) ellipse [x radius=2,  y radius=1.5]; 
	\draw[line width = 0.7, color=r2] (axis cs:15.9163,-0.633277) ellipse [x radius=2,  y radius=1.5]; 
	
	\end{axis}
	\end{tikzpicture}
	\caption{Potential function $G_{\P, \V^*}$ for finding new viewpoints (represented as red dots) after the second stage of the polygon generation shown in \figref{fig:accessible_area_voronoi}. The light red ellipses indicate the position of previous viewpoints. }
	\label{fig:accessible_area_sampling}	
\end{figure}

However, it is still possible that there are local maximizers of $G_{\P, \V^*}$ that are rather close to each other or, alternatively, between old viewpoints.
To prevent such accumulations, a subset of the candidates is selected that comply with a predefined clearance $\clearance$ in a second step.
The result is illustrated for an exemplary situation in \figref{fig:accessible_area_sampling} which shows the potential field of the polygon from \figref{fig:accessible_area_voronoi} (bottom left).
Here two new viewpoints are determined.


\begin{algorithm2e}[t]
	
	\newcommand\mycommfont[1]{\footnotesize\ttfamily\textcolor{k3}{#1}}
	\SetCommentSty{mycommfont}
	
%
%
	\SetKwFunction{algo}{CreatePolygon}
	\SetKwFunction{viewpoints}{GetViewpoints}
	\SetKwFunction{polygonfrom}{GeneratePolygonFrom}
	\SetKwProg{myalg}{}{}{}
	\KwIn{sytem state $s_0$, data $\D$}
	\Parameter{refinements $\refinements ,\,$clearance $\clearance,\,$ expansion $\expansion$}
	\hrulefill \\
	Set polygon $\P:=\{\}$ and viewpoints $\V:=\{s_0\}, \V^*:=\{\}$\;
	\For{$k\gets0$ \KwTo $\refinements$}
	{
		Set $i:=0$\;
		\ForEach{$\viewpoint \in \V$}
		{
			$i \leftarrow i + 1$\;
			$\P_i := \polygonfrom(\viewpoint,\,\D)$\;
			}
		$\P \leftarrow \text{unite}(\P, \P_1, \ldots, \P_i)$\;
		\If{$k \neq \refinements$}
		{
			$\V^* \leftarrow \V^* \bigcup \V$\;
			$\V \leftarrow \text{\viewpoints}(\P, \V^*)$\;
		}
	}

	\KwRet $\P$\;
	
	\SetKwProg{myproc}{Function}{}{}
	\myproc{\polygonfrom{$\viewpoint,\D$}}{
		
		$\P = \{\}$\;
		
		\ForEach{$S \in $ \emph{Sectors}}{
			Set $p_\text{nearest} :=$ nan, $d_\text{nearest} := \infty$\;
			\ForEach{$p \in \D$}{
				\If{$p \in S$ and $\dist(p, \viewpoint) < d_\text{nearest}$}{
					$p_\text{nearest} \leftarrow p$,
					$d_\text{nearest} \leftarrow \dist(p, \viewpoint)$\;
				}
			}
			\eIf{$p_\text{nearest} \neq$ nan}{
				Add $p_\text{nearest}$ to $\P$\;
				}
				{
				Add midpoint at end of sector to $\P$\;
				}
		}
		
		\KwRet $\P$\;}
	
	\SetKwProg{myproc}{Function}{}{}
	\myproc{\viewpoints{$\P, \V^*$}}{
	Set $\V \gets \{p \in \RR^2\,|\,p \text{ local maximizer of } G_{\P,\V^*}\}$\;
	\While{$M:=\{\,p \in \V\,|\, \dist(p,\,\V^*\bigcup[\V\setminus\{p\}]) < \clearance\} \neq \emptyset $}
	{
		$\V \gets \V \setminus \{p\}$ for one $p\in M$ \tcc*[r]{no accumulations}
	}
	\KwRet $\V$\;}
	\caption{Automated free-space polygon generation.}
	\label{alg:polygon}
\end{algorithm2e}

%% file: polygon_creation_a.pdf_tex
\begingroup%
  \makeatletter%
  \providecommand\color[2][]{%
    \errmessage{(Inkscape) Color is used for the text in Inkscape, but the package 'color.sty' is not loaded}%
    \renewcommand\color[2][]{}%
  }%
  \providecommand\transparent[1]{%
    \errmessage{(Inkscape) Transparency is used (non-zero) for the text in Inkscape, but the package 'transparent.sty' is not loaded}%
    \renewcommand\transparent[1]{}%
  }%
  \providecommand\rotatebox[2]{#2}%
  \newcommand*\fsize{\dimexpr\f@size pt\relax}%
  \newcommand*\lineheight[1]{\fontsize{\fsize}{#1\fsize}\selectfont}%
  \ifx\svgwidth\undefined%
    \setlength{\unitlength}{125.09861503bp}%
    \ifx\svgscale\undefined%
      \relax%
    \else%
      \setlength{\unitlength}{\unitlength * \real{\svgscale}}%
    \fi%
  \else%
    \setlength{\unitlength}{\svgwidth}%
  \fi%
  \global\let\svgwidth\undefined%
  \global\let\svgscale\undefined%
  \makeatother%
  \begin{picture}(1,1.019668)%
    \lineheight{1}%
    \setlength\tabcolsep{0pt}%
    \put(0,0){\includegraphics[width=\unitlength,page=1]{polygon_creation_a.pdf}}%
    \put(0.12278748,0.67946633){\color[rgb]{0,0,0}\makebox(0,0)[lt]{\lineheight{0.99599999}\smash{\begin{tabular}[t]{l}\small A\end{tabular}}}}%
    \put(0.41745003,0.46388197){\color[rgb]{0,0,0}\makebox(0,0)[lt]{\lineheight{0.99599999}\smash{\begin{tabular}[t]{l}\small B\end{tabular}}}}%
    \put(0.72948099,0.27096052){\color[rgb]{0,0,0}\makebox(0,0)[lt]{\lineheight{0.99599999}\smash{\begin{tabular}[t]{l}\small C\end{tabular}}}}%
    \put(0,0){\includegraphics[width=\unitlength,page=2]{polygon_creation_a.pdf}}%
    \put(0.0424759,0.88547575){\color[rgb]{0,0,0}\makebox(0,0)[lt]{\lineheight{0.99599999}\smash{\begin{tabular}[t]{l}\tiny\romannumber{1}\end{tabular}}}}%
    \put(0,0){\includegraphics[width=\unitlength,page=3]{polygon_creation_a.pdf}}%
    \put(0.4232535,0.76781858){\color[rgb]{0,0,0}\makebox(0,0)[lt]{\lineheight{0.99599999}\smash{\begin{tabular}[t]{l}\tiny\romannumber{2}\end{tabular}}}}%
    \put(0,0){\includegraphics[width=\unitlength,page=4]{polygon_creation_a.pdf}}%
    \put(0.72202343,0.48329593){\color[rgb]{0,0,0}\makebox(0,0)[lt]{\lineheight{0.99599999}\smash{\begin{tabular}[t]{l}\tiny\romannumber{3}\end{tabular}}}}%
    \put(0,0){\includegraphics[width=\unitlength,page=5]{polygon_creation_a.pdf}}%
    \put(0.88378135,0.04876803){\color[rgb]{0,0,0}\makebox(0,0)[lt]{\lineheight{0.99599999}\smash{\begin{tabular}[t]{l}\tiny\romannumber{4}\end{tabular}}}}%
    \put(0,0){\includegraphics[width=\unitlength,page=6]{polygon_creation_a.pdf}}%
    \put(8.20545796,0.5686689){\color[rgb]{0,0,0}\makebox(0,0)[lt]{\lineheight{0.99599999}\smash{\begin{tabular}[t]{l}A\end{tabular}}}}%
    \put(8.55857443,0.36207715){\color[rgb]{0,0,0}\makebox(0,0)[lt]{\lineheight{0.99599999}\smash{\begin{tabular}[t]{l}B\end{tabular}}}}%
    \put(8.78517286,0.31903743){\color[rgb]{0,0,0}\makebox(0,0)[lt]{\lineheight{0.99599999}\smash{\begin{tabular}[t]{l}C\end{tabular}}}}%
    \put(0,0){\includegraphics[width=\unitlength,page=7]{polygon_creation_a.pdf}}%
    \put(8.21057926,0.80165693){\color[rgb]{0,0,0}\makebox(0,0)[lt]{\lineheight{0.99599999}\smash{\begin{tabular}[t]{l}\romannumber{1}\end{tabular}}}}%
    \put(0,0){\includegraphics[width=\unitlength,page=8]{polygon_creation_a.pdf}}%
    \put(8.59135669,0.68399976){\color[rgb]{0,0,0}\makebox(0,0)[lt]{\lineheight{0.99599999}\smash{\begin{tabular}[t]{l}\romannumber{2}\end{tabular}}}}%
    \put(0,0){\includegraphics[width=\unitlength,page=9]{polygon_creation_a.pdf}}%
    \put(8.89012697,0.39947694){\color[rgb]{0,0,0}\makebox(0,0)[lt]{\lineheight{0.99599999}\smash{\begin{tabular}[t]{l}\romannumber{3}\end{tabular}}}}%
    \put(0,0){\includegraphics[width=\unitlength,page=10]{polygon_creation_a.pdf}}%
    \put(9.05188558,-0.03505087){\color[rgb]{0,0,0}\makebox(0,0)[lt]{\lineheight{0.99599999}\smash{\begin{tabular}[t]{l}\romannumber{4}\end{tabular}}}}%
    \put(0,0){\includegraphics[width=\unitlength,page=11]{polygon_creation_a.pdf}}%
  \end{picture}%
\endgroup%

%% file: polygon_creation_b.pdf_tex
\begingroup%
  \makeatletter%
  \providecommand\color[2][]{%
    \errmessage{(Inkscape) Color is used for the text in Inkscape, but the package 'color.sty' is not loaded}%
    \renewcommand\color[2][]{}%
  }%
  \providecommand\transparent[1]{%
    \errmessage{(Inkscape) Transparency is used (non-zero) for the text in Inkscape, but the package 'transparent.sty' is not loaded}%
    \renewcommand\transparent[1]{}%
  }%
  \providecommand\rotatebox[2]{#2}%
  \newcommand*\fsize{\dimexpr\f@size pt\relax}%
  \newcommand*\lineheight[1]{\fontsize{\fsize}{#1\fsize}\selectfont}%
  \ifx\svgwidth\undefined%
    \setlength{\unitlength}{125.09861503bp}%
    \ifx\svgscale\undefined%
      \relax%
    \else%
      \setlength{\unitlength}{\unitlength * \real{\svgscale}}%
    \fi%
  \else%
    \setlength{\unitlength}{\svgwidth}%
  \fi%
  \global\let\svgwidth\undefined%
  \global\let\svgscale\undefined%
  \makeatother%
  \begin{picture}(1,1.019668)%
    \lineheight{1}%
    \setlength\tabcolsep{0pt}%
    \put(0,0){\includegraphics[width=\unitlength,page=1]{polygon_creation_b.pdf}}%
    \put(-8.04531609,0.76328501){\color[rgb]{0,0,0}\makebox(0,0)[lt]{\lineheight{0.99599999}\smash{\begin{tabular}[t]{l}\small A\end{tabular}}}}%
    \put(-7.75065354,0.54770065){\color[rgb]{0,0,0}\makebox(0,0)[lt]{\lineheight{0.99599999}\smash{\begin{tabular}[t]{l}\small B\end{tabular}}}}%
    \put(-7.43862258,0.3547792){\color[rgb]{0,0,0}\makebox(0,0)[lt]{\lineheight{0.99599999}\smash{\begin{tabular}[t]{l}\small C\end{tabular}}}}%
    \put(0,0){\includegraphics[width=\unitlength,page=2]{polygon_creation_b.pdf}}%
    \put(-8.12562767,0.96929443){\color[rgb]{0,0,0}\makebox(0,0)[lt]{\lineheight{0.99599999}\smash{\begin{tabular}[t]{l}\tiny\romannumber{1}\end{tabular}}}}%
    \put(0,0){\includegraphics[width=\unitlength,page=3]{polygon_creation_b.pdf}}%
    \put(-7.74485007,0.85163726){\color[rgb]{0,0,0}\makebox(0,0)[lt]{\lineheight{0.99599999}\smash{\begin{tabular}[t]{l}\tiny\romannumber{2}\end{tabular}}}}%
    \put(0,0){\includegraphics[width=\unitlength,page=4]{polygon_creation_b.pdf}}%
    \put(-7.44608014,0.56711461){\color[rgb]{0,0,0}\makebox(0,0)[lt]{\lineheight{0.99599999}\smash{\begin{tabular}[t]{l}\tiny\romannumber{3}\end{tabular}}}}%
    \put(0,0){\includegraphics[width=\unitlength,page=5]{polygon_creation_b.pdf}}%
    \put(-7.28432222,0.13258672){\color[rgb]{0,0,0}\makebox(0,0)[lt]{\lineheight{0.99599999}\smash{\begin{tabular}[t]{l}\tiny\romannumber{4}\end{tabular}}}}%
    \put(0,0){\includegraphics[width=\unitlength,page=6]{polygon_creation_b.pdf}}%
    \put(0.03735439,0.65248759){\color[rgb]{0,0,0}\makebox(0,0)[lt]{\lineheight{0.99599999}\smash{\begin{tabular}[t]{l}\small A\end{tabular}}}}%
    \put(0.39047086,0.44589583){\color[rgb]{0,0,0}\makebox(0,0)[lt]{\lineheight{0.99599999}\smash{\begin{tabular}[t]{l}\small B\end{tabular}}}}%
    \put(0.61706929,0.40285612){\color[rgb]{0,0,0}\makebox(0,0)[lt]{\lineheight{0.99599999}\smash{\begin{tabular}[t]{l}\small C\end{tabular}}}}%
    \put(0,0){\includegraphics[width=\unitlength,page=7]{polygon_creation_b.pdf}}%
    \put(0.04247569,0.88547561){\color[rgb]{0,0,0}\makebox(0,0)[lt]{\lineheight{0.99599999}\smash{\begin{tabular}[t]{l}\tiny\romannumber{1}\end{tabular}}}}%
    \put(0,0){\includegraphics[width=\unitlength,page=8]{polygon_creation_b.pdf}}%
    \put(0.42325312,0.76781844){\color[rgb]{0,0,0}\makebox(0,0)[lt]{\lineheight{0.99599999}\smash{\begin{tabular}[t]{l}\tiny\romannumber{2}\end{tabular}}}}%
    \put(0,0){\includegraphics[width=\unitlength,page=9]{polygon_creation_b.pdf}}%
    \put(0.72202339,0.48329562){\color[rgb]{0,0,0}\makebox(0,0)[lt]{\lineheight{0.99599999}\smash{\begin{tabular}[t]{l}\tiny\romannumber{3}\end{tabular}}}}%
    \put(0,0){\includegraphics[width=\unitlength,page=10]{polygon_creation_b.pdf}}%
    \put(0.88378201,0.04876782){\color[rgb]{0,0,0}\makebox(0,0)[lt]{\lineheight{0.99599999}\smash{\begin{tabular}[t]{l}\tiny\romannumber{4}\end{tabular}}}}%
    \put(0,0){\includegraphics[width=\unitlength,page=11]{polygon_creation_b.pdf}}%
  \end{picture}%
\endgroup%

%% file: sections/HSAS.tex
An essential feature of self-driving cars is the ability to rapidly plan a set of executable actions even in complex situations. A well-known approach that fulfills these requirements in static settings is the hybrid-state \astar algorithm by \cite{dolgov08gppSTAIR}. Building upon its idea of hybrid nodes, we present details of an extension that is able to also capture time-varying parts of the vehicles environment.

\subsection{Kinematics} \label{sec:hsas_kinematics}
To consider moving objects $\O$ in a vehicle's planned path, it is necessary to take the ego-speed into account and to allow changes to it. The simplest kinematic description incorporating this is the single-track model \citep[e.g.][]{Luca1998NonholonomicCar} given as
\begin{equation}
\begin{array}{c}
\begin{pmatrix}
\dot{x},
\dot{y},
\dot{\psi},
\dot{v}
\end{pmatrix}^\top = \begin{pmatrix}
v\cos(\psi),
v\sin(\psi),
\nicefrac{v}{L} \tan(\beta),
a
\end{pmatrix}^\top.
\end{array}
\label{eq:single-track-model-simple}
\end{equation}

Therein, the car's state consists of the $(x,y)$-position (defined as the center of the rear axle), the orientation $\psi$ as well as the speed $v$. The wheelbase $L$ denotes a vehicle-specific parameter. The motion of the car may be influenced by the acceleration $a$ or by the steering angle $\beta$.

Like in the ordinary hybrid-state \astar algorithm, the motion of the car is regarded by applying discrete controls, in our case $\A:=\{a_1, \ldots, a_{N_a}\}$ and $\B:=\{\beta_1, \ldots, \beta_{N_\beta}\}$. This results in a continuous state representation $(t, x, y, \psi, v)$, with $t$ denoting the time (in order to take varying locations of obstacles into account). The \astar search space is then formed by discretization of these state variables, whereby the continuous representation and its corresponding discrete cell are stored jointly.
After termination of the search algorithm, a feasible path in the sense of the kinematic model \eqref{eq:single-track-model-simple} is available by following the continuous representation of the corresponding nodes.

\subsection{Dynamic Voronoi Field} \label{subsec:voronoi}
As proposed in the original hybrid-state \astar method, our approach is guided by a cost function based on a Voronoi diagram. During planning, its edges serve as a  reference path to avoid collisions. Moreover, a Voronoi diagram for a free-space polygon can be computed very efficiently by the Sweepline algorithm \citep[e.g.][]{Sydorchuk2012voronoi}, which allows the input sites to be solely in the form of segments. The resulting set of edges, which come typically in the form of curves, are then approximated by straight lines to enable the fast computation of distances to them. 

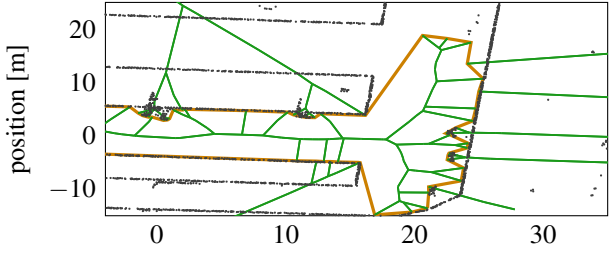
\begin{figure}[t]
	
	\newcommand{\figwidth}{0.45}
	\newcommand{\figheight}{0.18}
	\newcommand{\xmin}{-4}
	\newcommand{\xmax}{35}
	\newcommand{\ymin}{-15}
	\newcommand{\ymax}{25}
	\newcommand{\datamarksize}{0.2}
	\newcommand{\polylinewidth}{1.2}
	\newcommand{\voronoilinewidth}{0.7}
	
	\pgfplotsset{xtick style={draw=none}, ytick style={draw=none}}
	
	\newcommand\noblankfileline{\expandafter\noblankaux\fileline\relax}
	\def\noblankaux#1 \relax{#1}
	
	\centering
	\begin{tikzpicture}
	\begin{axis}[
	width=\figwidth\textwidth,
	height=\figheight\textheight,
	ylabel= {position [\si{\meter}]},
	xmin = \xmin,
	xmax = \xmax,
	ymin = \ymin,
	ymax = \ymax,
	]
	
	\addplot[each nth point=2, filter discard warning=false, unbounded coords=discard, only marks, mark size = \datamarksize, color=\datacolor] table [col sep=comma] {data/voronoi_filter/results_obstacles.dat};
	
	\addplot[unbounded coords=discard, line width = \polylinewidth, color=\polycolor] table [col sep=comma] {data/voronoi_filter/results_poly.dat};		
	
	\PlotVoronoi{data/}{voronoi_filter/results_voronoi_path_filelist.dat}		
	
	\end{axis}	
	\end{tikzpicture}	
	\caption{(Unfiltered) Voronoi diagram (green lines) for the second free-space polygon (orange lines) from \figref{fig:accessible_area_voronoi}.}
	\label{fig:voronoi_filter}
\end{figure}

The resulting Voronoi diagram usually contains edges that are not suitable as references for path planning., e.g., those that lie outside or close to the border of the polygon, as shown in \figref{fig:voronoi_filter}. However, the corresponding vertices can be eliminated easily, so that the Voronoi reference paths $\R$ presented in \figref{fig:accessible_area_voronoi} remain. Note that the small branch-offs shown could be filtered out further, but have almost no influence on the planning result in practice, as they usually do not comply with the vehicles kinematics \eqref{eq:single-track-model-simple}.

In cases where other dynamic objects $\O$ are \textit{within} the current free-space polygon (at least partially), Voronoi reference paths $\R_t$ are computed for a number of discrete times to take their movement into account as well. This is easy to parallelize and can therefore be conducted very efficiently. The result is then used to define the time-dependent Voronoi field
\begin{align}
\begin{split}
\varrho_{\R_t, \P, \O_t}(s_t) := &\left(\frac{\alpha}{\alpha+\dist^{3}_{\P,\O_t}(s_t)}\right) \left(\frac{\dist^2_{R_t}(s_t)}{\dist^3_{\P,\O_t}(s_t)+ \dist^2_{\R_t}(s_t)}\right)\\
&\left(\frac{\dist^3_{\P,\O_t}(s_t)-\dist_{\max}}{\dist_{\max}}\right)^2
\label{eq:voronoi_potential}
\end{split}
\end{align}
for the vehicle state $s_t$ at the discrete time $t$ and with $\alpha, \dist_{\max}\in \RR_{>0}$. Further, $\dist^2_{\R_t}(\cdot)$ denotes the distance from the vehicle's reference point $(x_t,y_t)$ to the Voronoi path. Moreover, $\dist^{3}_{\P,\O_t}(\cdot)$ represents the minimal distance of the ego-vehicle to the polygon $\P$ and all dynamic obstacles $\O_t$, dependent on $(x_t, y_t, \psi_t)$.

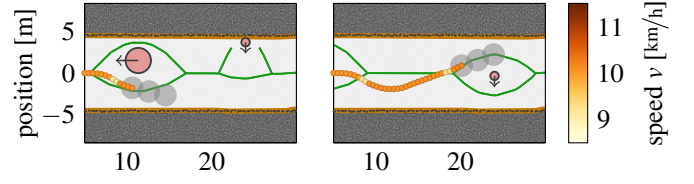
\begin{figure}[t]
	
	\pgfplotsset{xtick style={draw=none}, ytick style={draw=none}}
	
	\newcommand{\figwidth}{0.24}
	\newcommand{\figheight}{0.14}
	\newcommand{\xmin}{5}
	\newcommand{\xmax}{30}
	\newcommand{\ymin}{-8}
	\newcommand{\ymax}{8}
	\newcommand{\vmin}{8.5}
	\newcommand{\vmax}{11.5}
	\newcommand{\objectcolor}{r2}
	\newcommand{\speedcolor}{k2}
	\newcommand{\vehiclecolor}{k3}
	\newcommand{\polylinewidth}{2}
	\newcommand{\voronoilinewidth}{0.7}
	\newcommand{\objectlinewidth}{0.7}
	\newcommand{\speedlinewidth}{0.6}
	\newcommand{\vehiclelinewidth}{0.9}
	\newcommand{\datamarksize}{0.2}
	\newcommand{\speedscale}{1.2}
	
	\newcommand\noblankfileline{\expandafter\noblankaux\fileline\relax}
	\def\noblankaux#1 \relax{#1}
	
	\centering
	\begin{tikzpicture}
	\begin{axis}[
	width=\figwidth\textwidth,
	height=\figheight\textheight,
	axis equal,
	ylabel= {position [\si{\meter}]},
	xlabel= \empty, 
	ylabel shift = -2 mm,
	xmin = \xmin,
	xmax = \xmax,
	ymin = \ymin,
	ymax = \ymax,
	xtick = {10,20},
	colormap/YlOrBr,
	view={0}{90},
	]
	
	\PlotSituation{data/voronoi/}{results_obstacles.dat}{results_poly.dat}{\xmin}{\xmax}{\ymin}{\ymax}
	
	\PlotVoronoi{data/}{voronoi/results_voronoi_path_13_filelist.dat}
	
	\PlotTrajectorySearch{data/voronoi/}{results_trajectory_13.dat}{results_problem_info.dat}{\vmin}{\vmax}
	
	\PlotMovingObjectZero{data/voronoi/objects/results_moving_obstacle_files}{14}
	
	\PlotMovingObjectOne{data/voronoi/objects/results_moving_obstacle_files}{14}	
	
	\end{axis}
	\end{tikzpicture}
	\hfill
	\begin{tikzpicture}
	\begin{axis}[
	width=\figwidth\textwidth,
	height=\figheight\textheight,
	axis equal,
	xlabel= \empty,
	xmin = \xmin,
	xmax = \xmax,
	ymin = \ymin,
	ymax = \ymax,
	ytick = \empty,
	xtick = {10,20},
	colormap/YlOrBr,
	colorbar,
	colorbar/width=2.5mm,
	colorbar style={
		ylabel=speed $v$ [\si{\kilo\metre\per\hour}],
		xshift=-0mm,
	},
	view={0}{90},
	]
	
	\PlotSituation{data/voronoi/}{results_obstacles.dat}{results_poly.dat}{\xmin}{\xmax}{\ymin}{\ymax}
	
	\PlotVoronoi{data/}{voronoi/results_voronoi_path_25_filelist.dat}
	
	\PlotTrajectorySearch{data/voronoi/}{results_trajectory_25.dat}{results_problem_info.dat}{\vmin}{\vmax}
	
	\PlotMovingObjectZero{data/voronoi/objects/results_moving_obstacle_files}{26}
	
	\PlotMovingObjectOne{data/voronoi/objects/results_moving_obstacle_files}{26}
	
	\end{axis}
	\end{tikzpicture}
	\caption{Avoiding moving obstacles (represented as pale red circles) using the dynamic Voronoi path (green lines). The planning result is shown by the sequences of colored dots at two different times. The orange lines show the free-space polygon and the three gray circles are used to check for collisions of the ego-vehicle.}
	\label{fig:dynamic_voronoi_example}
\end{figure}

The definition in \eqref{eq:voronoi_potential} has similar characteristics as the potential introduced by \cite{dolgov08gppSTAIR}, i.e. $\varrho_{\R_t, \P, \O_t} \in [0,1]$ and its property support the navigation in narrow areas. However, it is a generalization in the sense that it considers time-varying information and performs full collision checks. An example of a maneuver based on this is given in \figref{fig:dynamic_voronoi_example}.

\subsection{Costs and Heuristics}
Given a desired goal state $s_T$, the movement cost $g$ to a node $s_t$ in our proposed \astar search penalizes three things: 
\begin{enumerate}[label=(\roman*)]
	\item Deviations from a desired speed $v_\text{set}$: \[g_v := \frac{(v_t-v_\text{set})^2}{\max(v_\text{set}^2, v_\text{set,\,min}^2)}.\]
	\item Discounted proximity to obstacles: \[g_o := \varrho_{\R_t, \P, \O_t}(s_t) \frac{p_\circledcirc}{p_\oplus}.\]
	\item Making a step: \[g_p:=1.\]
\end{enumerate}
The lower limit $v_\text{set,\,min}$ for the reference speed is introduced to avoid singularities at small values (e.g. when stopping the car). The Voronoi field $\varrho_{\R_t, \P, \O_t}$ is discounted based on the expected remaining relative path length. This is obtained as the ratio of $p_\circledcirc$ and $p_\oplus$, which represent estimates of the absolute path lengths to the goal node $s_T$ from the current state $s_t$ and from the initial node, respectively. To approximate these two quantities, we employ the obstacle-free shortest path proposed by \cite{Dubins1957OnCO} (for the case of driving forward) or \cite{reeds1990} (if driving backwards is also allowed, e.g. during parking).
Finally, all three components are normalized with the relative stepsize $\nicefrac{p_{\ominus}}{p_\oplus}$, which can be assumed fixed for a constant discretization of the \astar search space. 
Altogether, the costs $g$ are then defined as the weighted sum 
\begin{equation*}
	g:=(w_vg_v + w_og_o + w_p g_p) \frac{p_\ominus}{p_\oplus}, \quad w_v,w_o,w_p \in \RR_{\geq 0}.
\end{equation*}
The \astar heuristic $h$, on the other hand, should estimate the total costs to reach the  goal $s_T$ from a given node. A very optimistic and certainly admissible lower bound for this would be the expected remaining (obstacle-free) path length $\nicefrac{p_\circledcirc}{p_\oplus}$ for the assumption of having no further deviations from both the reference speed and the Voronoi path. Nevertheless, we propose a more pessimistic approach by expecting the current deviations to persist, hence resulting in the heuristic
\begin{equation*}
h:=(w_vg_v + w_og_o + w_p g_p) \frac{p_\circledcirc}{p_\oplus}.
\end{equation*}
%
In practice, using $h$ typically provides good approximations of the true costs and thus results in fast convergence of the planner. However, this \astar algorithm is not admissible.

%% file: sections/Optimization.tex
The preceding sections describe how model-based trajectories can be planned in complex and dynamic environments. However, when it comes to the actual control of an autonomous vehicle, additional constraints and optimality criteria may need to be applied to determine appropriate control signals. For example, it makes sense to consider continuous controls whose rate of change is bounded, to meet the comfort requirements of a passenger. In addition, physical aspects, such as the propagation time of signals, should be considered to achieve high precision in the maneuvers performed.

Model predictive control represent a framework for meeting the above-mentioned demands \citep[e.g.][]{Rick2019}. Therein, optimal control problems of the form
\begin{align}
\begin{split}
\min_{z,u,T} \quad &J(z,u,T)\\
\text{s.t. } \dot{z}(t) &= f(z(t),u(t)),\\
z_\text{min} &\leq z(t) \leq z_\text{max},\\
u_\text{min} &\leq u(t) \leq u_\text{max},\\
\varPsi(z(0)) &= 0,\\
C(z(t),u(t),t) &\leq 0 \text{ for all } t \in [0,T]
\end{split}
\label{eq:ocp}
\tag{OCP}
\end{align}
are solved to find states $z\in\C^1([0,T],\RR^{n_z})$, controls $u\in\C^0([0,T],\RR^{n_u})$ and the process time $T\in\RR_{>0}$, that optimize an objective function $J$, at high frequency. Furthermore, the system's dynamic $f$ is considered and the optimization variables are subject to box constraints, initial conditions $\varPsi$ and path constraints $C$. The main advantage of this problem formulation lies in its universality, which allows the application for various different driving maneuvers.

In the following, we provide details of an optimal control problem formulation which allows to compute sophisticated trajectories, based on the planning results of the time-dependent hybrid-state \astar algorithm, with the goal of controlling a real-world vehicle (see also \secref{subsec:vehicle}).

\subsection{Delayed Single-Track Model}
To be applicable to our research car, the dynamic vehicle \mbox{model $f$} should include the propagation time for applying the steering angle $\beta$ and the acceleration signal $a$, which are approximately \SI{120}{\milli\second} and \SI{500}{\milli\second} respectively.
Furthermore, the representation of the dynamics should be kept simple, to enable fast updates in the context of model predictive control.
Therefore, we propose the following single-track model
%
\begin{equation}
\begin{aligned}
\dot{x} 	&= v \cos(\psi), \qquad 				& \dot{v} 				&= \tilde{a},\qquad & \dot{\tilde{\beta}}   &= \nicefrac{(\beta-\tilde{\beta})}{\Delta t_\beta},\\
\dot{y} 	&= v \sin(\psi),        				& \dot{\beta} 			&= \omega_\beta, 	& \dot{\tilde{a}} 		&= \nicefrac{(a-\tilde{a})}{\Delta t_a}, \\
\dot{\psi}  &= \nicefrac{v}{L}\tan(\tilde{\beta}), 	& \dot{a} 				&= j, 				& \\
\end{aligned}
\label{eq:model_delay}
\end{equation}
which extends the kinematics in \secref{sec:hsas_kinematics} by introducing a delayed acceleration $\tilde{a}$ and steering angle $\tilde{\beta}$ modeled as \mbox{PT1-elements}.
The signal propagation time is then incorporated (indirectly) by means of the time lags $\Delta t_a$ and $\Delta t_\beta$.
Moreover, changes in the control values can be bounded by considering the jerk $j$ and the steering angle velocity $\omega_\beta$, respectively.

\subsection{Objective Function}
In order to take safety and comfort aspects into account in the solution of \eqref{eq:ocp}, the objective function $J$ weights different terms against each other.
First, large changes in the acceleration and steering angle are avoided by 
%
\begin{align*}
	J_\text{controls} :=
	w_0 \int_{0}^{T} j(t) \d t +
	w_1 \int_{0}^{T} \omega_\beta(t) \d t, \quad w_0, w_1 \in \RR_{>0}.
\end{align*}
In addition, deviations from the goal speed $v_\text{set}$ is penalized by
\begin{align*}
	J_\text{states} :=
	w_2 \int_{0}^{T} (v(t) - v_\text{set}) \d t, \quad w_2 \in \RR_{>0}.
\end{align*}
Finally, the goal position is also incorporated within the objective function instead of considering it as a hard constraint. This enhances the flexibility of the optimization and results in
%
\begin{align*}
	J_\text{goal} :=
	w_3 \left( (x(T) - x_\text{goal})^2 + (y(T) - y_\text{goal})^2 \right), \quad w_3 \in \RR_{>0}.
\end{align*}
Altogether, the objective function is defined as
\begin{align*}
	J := J_\text{controls} + J_\text{states} + J_\text{goal}.
\end{align*}
The choice of weighting does not only influence the driving behavior but also has a great impact on the convergence rate and robustness of the optimization.

\subsection{Constraints \& Initial Guess}

The latest estimated state of the vehicle defines the initial condition considered in $\varPsi$ of \eqref{eq:ocp}.
Because the final state is already regarded by the objective function, no further conditions for the final state at time $T$ are required.

To prevent the vehicle from colliding with obstacles in terms of the path constraints $C$, the car is approximated by overlapping circles, see also \figref{fig:dynamic_voronoi_example}. In particular, static obstacles are taken into account by enforcing to stay within a free-space polygon. Note that the latter can be reused from the motion planner. Furthermore, in contrast to the results presented by \cite{Sommer2018icatt}, moving road users are considered in the path constraints as well.


Since the methods for solving optimal control problems usually only find local minima, a good initial guess is required. In the setting of model predictive control, typically the last solution can be used for this. However, if there is a strong change in the goal state provided by the decision making, a complete reoptimization might be necessary. In this case there are several ways to obtain an initial estimate, e.g. by linear interpolation, computing a Reeds-Shepp path or by using the trajectory provided by the time-dependent hybrid-state \astar algorithm. We compare the last two options in \secref{subsec:result:ocp}.

%% file: sections/Results.tex

We provide numerical results for model-based planning and optimal control in simulated situations, which should exemplarily cover a wide range of requirements on the solver. Thereby, the environment will be represented as a noisy point cloud. Moving obstacles are assumed to be approximated by circles. We restrict their motion to be linear in all cases to allow comparability between different situations. However, all methods are applicable to arbitrary estimates of their movement. The experiments are conducted on a Intel i7-4790 with \SI{3.6}{\giga\hertz}. A summary of all hyperparameters used can be found in \appref{app:params}.

\subsection{Path Planning}

Given a certain situation, the motion planner performs two preliminary steps: it generates a free-space polygon (see \secref{sec:polygon}) and precomputes the corresponding Voronoi reference paths for the next \SI{10}{\second} (see \secref{subsec:voronoi}). A goal is then determined based on the static Voronoi path in accordance with the requirements of a superordinate planner (e.g. \emph{turn left} or \emph{park}). During planning, the costs and heuristics of the time-dependent hybrid-state \astar algorithm are evaluated very efficiently by precalculating the Reeds-Shepp or Dubins paths between discrete states. In addition, the distances to the polygon and reference path are memorized when evaluating the Voronoi potential $\varrho_{\R_t, \P, \O_t}$. On the one hand, this typically reduces the computation time for individual problems by half. On the other hand, these intermediate results can also be used for efficient decision making, which requires multiple evaluations of the same scene with different goal nodes in order to analyze possible actions. Finally, the specific implementation of the \astar's open set combines the advantages of a hashmap and a priority queue, providing fast content checking and access.

\input{sections/result_plots}

Solutions for exemplary static and dynamic scenarios are illustrated in \figref{fig:results}. The corresponding number of expanded nodes and computation times are presented in \tabref{tab:results}. In general, the \astar expansion is strongly oriented towards the Voronoi reference path. This leads to very safe maneuvers on the one hand and to a small number of expansions away from the final solution on the other. This also applies to complicated tasks such as parking into a narrow parking space or turning with several moving objects.

The overall computation time results from three components. While the calculation of the free-space polygon is approximately constant for all situations, the creation of the Voronoi path depends on the presence of moving obstacles. If this is the case, dynamic reference paths are generated based on the time discretization. Note that these two auxiliary quantities only have to be determined once for a common scene with different goals. Finally, the execution of the time-dependent hybrid-state \astar then terminates for simple situations in less than \SI{5}{\milli\second}. Its computation time, however, increases with more distant goals, such as when turning right, or when more expansion is required in difficult scenarios. Even in the most demanding cases, solutions are found in less than \SI{40}{\milli\second}, which makes this method a fast, accurate and robust basis for tactical decisions in urban environments.

\begin{table}[b]
	\newcommand{\figcellcolor}{\cellcolor{k4}}
	\captionsetup{width=0.8\linewidth} 
	\centering
	\caption{Number of expanded nodes and computation times to solve the situations shown in \figref{fig:results}. [$\cdot$]-brackets: coarser discretization of \SI{1}{\meter} for comparison (instead of \SI{0.5}{\meter}). $\dagger$: 3 steps for polygon creation (instead of 2). }
	\label{tab:results}
	\begin{tabular}{p{0.1mm}c!{\vrule width 1pt}c|c!{\vrule width 1pt}c|c|c}
		\multicolumn{2}{c!{\vrule width 1pt}}{} & \multicolumn{2}{c!{\vrule width 1pt}}{\textbf{Exp. Nodes}} & \multicolumn{3}{c}{\textbf{Computation Time} [\si{\milli\second}]}  \\
		\multicolumn{2}{c!{\vrule width 1pt}}{\textbf{Task}} & \textbf{Opened} & \textbf{Closed} & \textbf{Polygon} & \textbf{Voronoi} & \textbf{Hybrid \astar} \\
		\Xhline{1pt}
		
		\figcellcolor&Turn & 5106 & 472 & \multirow{2}{*}{3.0} & \multirow{2}{*}{0.96} & 4.4 \\
		\figcellcolor&Left 				& [4019] & [323] & & & [2.9] \\\cline{2-2}
		\figcellcolor&Turn & 22282 & 2618 & \multirow{2}{*}{$2.5^\dagger$} & \multirow{2}{*}{0.81} & 20.2 \\
		\figcellcolor&Right							& [12284] & [1319] &  &  & [10.4] \\\cline{2-2}	
		\multirow{-5}{*}{\figcellcolor\hspace{-1mm}\rotatebox[origin=c]{90}{\figref{fig:results_a} }}&Parking & 21989 & 2531 & 3.1 & 1.1 & 38.5 \\
		\Xhline{1pt}
		\multicolumn{2}{c!{\vrule width 1pt}}{\figref{fig:results_b}} & 3604 & 335 & 2.9 & 15.5 & 3.5 \\\cline{1-2}
		\multicolumn{2}{c!{\vrule width 1pt}}{\figref{fig:results_c}} & 24918 & 2847 & 2.9 & 21.4 & 28.6 \\	
	\end{tabular}
\end{table}

\subsection{Optimal Control}
\label{subsec:result:ocp}

Based on the result of the path planning, the optimal control problem from \secref{sec:optimization} is solved to compute executable control signals for a self-driving car. In general, this can be done efficiently by transcribing the formulation \eqref{eq:ocp} into a nonlinear optimization problem (NLP) \citep[e.g.][]{knauer2019}. The resulting task is, in turn, typically characterized by very sparse Jacobian and Hessian matrices which are exploited by highly advanced NLP solvers like \worhp \citep{buskens2012esa}. 

\figref{fig:ocp_results} illustrates the results of this optimization step for the turning maneuver with obstacles shown in \figref{fig:results_c}. For its computation, the existing free-space polygon is reused and WORHP's sequential quadratic programming method (SQP) is applied. One can see, that the optimized turning maneuver deviates from the planned reference solution in its path and being a few seconds shorter. The control values are smooth and especially the steering commands are very targeted if necessary and minimal otherwise.

\input{sections/results_ocp_plots}

Although this optimization task is rather complex, the solution shown can be found after 8 SQP steps within \SI{290}{\milli\second} on the basis of its sophisticated initial estimate. Looking at the same problem without the moving obstacles, as when turning left in \figref{fig:results_a}, even only 4 SQP steps and 60 ms are necessary. The benefit of reusing the path planning result as an approximation for the solution of the optimization step becomes particularly clear when comparing it with the Reeds-Shepp path (see \figref{fig:ocp_results}). If the latter is used as an initial guess, the computation effort increases to 6 SQP steps and \SI{120}{\milli\second} for the case without moving obstacles. In the corresponding dynamic scenario, at best insufficient local minima can be identified. Thus, the proposed method leads not only to reduced processing times, but most importantly to an improved robustness.

%% file: sections/result_plots.tex
\begin{figure}
	
	\newcommand{\subcaptionshiftupwards}{-3mm}
	\newcommand{\subcaptionshiftdownwards}{2mm}
	
	\newcommand{\shiftedsubcaption}[1]{
		\vspace*{\subcaptionshiftupwards}
		\subcaption{#1}
		\vspace*{\subcaptionshiftdownwards}
	}
	
	\newcommand{\lastshiftedsubcaption}[1]{
		\vspace*{\subcaptionshiftupwards}
		\subcaption{#1}
	}
		
	\pgfplotsset{xtick style={draw=none}, ytick style={draw=none}}
	
	\newcommand{\figwidth}{0.46}
	\newcommand{\figheight}{0.14}

	\newcommand{\branchcolor}{c1}
	\newcommand{\objectcolor}{r2}
	\newcommand{\speedcolor}{k2}
	\newcommand{\vehiclecolor}{k2}
	\newcommand{\polylinewidth}{2}
	\newcommand{\voronoilinewidth}{0.3}
	\newcommand{\branchlinewidth}{0.2}
	\newcommand{\objectlinewidth}{0.7}
	\newcommand{\speedlinewidth}{0.6}
	\newcommand{\vehiclelinewidth}{0.5}
	\newcommand{\datamarksize}{0.2}
	\newcommand{\speedscale}{1.2}
	
	\newcommand{\textboxcolor}{k4}
	
	\newcommand{\vmin}{-5}
	\newcommand{\vmax}{25}
	
	\newcommand\noblankfileline{\expandafter\noblankaux\fileline\relax}
	\def\noblankaux#1 \relax{#1}	
	
	\flushleft
	\hspace{-1.2em}
	\begin{minipage}[c]{0.93\linewidth}
	\begin{subfigure}[t]{1.0\textwidth}
	\begin{tikzpicture}
	
		\newcommand{\xmin}{-3}
		\newcommand{\xmax}{25}
		\newcommand{\ymin}{-2}
		\newcommand{\ymax}{18}
		
		\begin{axis}[
		width=\figwidth\textwidth,
		height=\figheight\textheight,
		xmin = \xmin,
		xmax = \xmax,
		ymin = \ymin,
		ymax = \ymax,
		ytick = {10},
		yticklabel = {\phantom{10}},
		axis equal,
		colormap/YlOrBr,
		view={0}{90},
		]
		
		\PlotSituation{data/results/planning/turning/}{results_obstacles.dat}{results_poly.dat}{\xmin}{\xmax}{\ymin}{\ymax}
		
		\PlotVoronoi{data/}{results/planning/turning/results_voronoi_path_filelist.dat}		
		
		\PlotTrajectorySearchBranches{data/results/planning/turning/}{results_branch_files.dat}
				
		\PlotTrajectorySearch{data/results/planning/turning/}{results_trajectory_0.dat}{results_problem_info.dat}{\vmin}{\vmax}
		
		\node[draw, fill=\textboxcolor, anchor=north west] at (axis cs: \xmin,19) {\tiny $\SI{20}{\kilo\meter\per\hour}$};
		
		\end{axis}
	\end{tikzpicture}
	\begin{tikzpicture}
	
		\newcommand{\xmin}{-3}
		\newcommand{\xmax}{35}
		\newcommand{\ymin}{-5}
		\newcommand{\ymax}{17}
			
		\begin{axis}[
		width=\figwidth\textwidth,
		height=\figheight\textheight,
		xmin = \xmin,
		xmax = \xmax,
		ymin = \ymin,
		ymax = \ymax,
		ytick = \empty,
		axis equal,
		colormap/YlOrBr,
		view={0}{90},
		]
		
		\PlotSituation{data/results/planning/scurveturn/}{results_obstacles.dat}{results_poly.dat}{\xmin}{\xmax}{\ymin}{\ymax}
		
		\PlotVoronoi{data/}{results/planning/scurveturn/results_voronoi_path_filelist.dat}	
		
		\PlotTrajectorySearchBranches{data/results/planning/scurveturn/}{results_branch_files.dat}
				
		\PlotTrajectorySearch{data/results/planning/scurveturn/}{results_trajectory_0.dat}{results_problem_info.dat}{\vmin}{\vmax}			
		\node[draw, fill=\textboxcolor, anchor=north west] at (axis cs: \xmin,21) {\tiny $\SI{20}{\kilo\meter\per\hour}$};
		
		\end{axis}
	\end{tikzpicture}	
	\begin{tikzpicture}
	
		\newcommand{\xmin}{-3}
		\newcommand{\xmax}{25}
		\newcommand{\ymin}{-12}
		\newcommand{\ymax}{5}
	
		\begin{axis}[
		width=\figwidth\textwidth,
		height=\figheight\textheight,
		xmin = \xmin,
		xmax = \xmax,
		ymin = \ymin,
		ymax = \ymax,
		ytick = \empty,
		axis equal,
		colormap/YlOrBr,
		view={0}{90},
		]
		
		\PlotSituation{data/results/planning/parking/}{results_obstacles.dat}{results_poly.dat}{\xmin}{\xmax}{\ymin}{\ymax}
		
		\PlotVoronoi{data/}{results/planning/parking/results_voronoi_path_filelist.dat}
		
		\PlotTrajectorySearchBranches{data/results/planning/parking/}{results_branch_files.dat}		
		
		\PlotTrajectorySearch{data/results/planning/parking/}{results_trajectory.dat}{results_problem_info.dat}{\vmin}{\vmax}				
		
		\node[draw, fill=\textboxcolor, anchor=north west] at (axis cs: \xmin,-9) {\tiny $\SI{0}{\kilo\meter\per\hour}$};
		
		\end{axis}
	\end{tikzpicture}
	\shiftedsubcaption{Static planning: Turning left; Turning right; Parking.}
	\label{fig:results_a}
	\end{subfigure}
	\begin{subfigure}{1.0\textwidth}
		
	\newcommand{\xmin}{-3}
	\newcommand{\xmax}{30}
	\newcommand{\ymin}{-8}
	\newcommand{\ymax}{8}
	
	\begin{tikzpicture}
		\begin{axis}[
		width=\figwidth\textwidth,
		height=\figheight\textheight,
		xmin = \xmin,
		xmax = \xmax,
		ymin = \ymin,
		ymax = \ymax,
		ytick = {0,10},
		axis equal,
		colormap/YlOrBr,
		view={0}{90},
		]
		
		\PlotSituation{data/results/planning/overtaking/}{results_obstacles.dat}{results_poly.dat}{\xmin}{\xmax}{\ymin}{\ymax}
		
		\PlotVoronoi{data/}{results/planning/overtaking/results_voronoi_path_3_filelist.dat}
		
		\PlotTrajectorySearchBranches{data/results/planning/overtaking/}{results_branch_files.dat}
		
		\PlotTrajectorySearch{data/results/planning/overtaking/}{results_trajectory_0.dat}{results_problem_info.dat}{\vmin}{\vmax}
		
		\PlotMovingObjectZero{data/results/planning/overtaking/objects/results_moving_obstacle_files}{4}		
		
		\node[draw, fill=\textboxcolor, anchor=north west] at (axis cs: \xmin,13) {\tiny $\SI{20}{\kilo\meter\per\hour}$};
		
		\end{axis}
	\end{tikzpicture}
	\begin{tikzpicture}
		\begin{axis}[
		width=\figwidth\textwidth,
		height=\figheight\textheight,
		xmin = \xmin,
		xmax = \xmax,
		ymin = \ymin,
		ymax = \ymax,
		ytick = \empty,
		axis equal,
		colormap/YlOrBr,
		view={0}{90},
		]
		
		\PlotSituation{data/results/planning/overtaking/}{results_obstacles.dat}{results_poly.dat}{\xmin}{\xmax}{\ymin}{\ymax}
		
		\PlotVoronoi{data/}{results/planning/overtaking/results_voronoi_path_8_filelist.dat}
		
		\PlotTrajectorySearchBranches{data/results/planning/overtaking/}{results_branch_files.dat}			
		
		\PlotTrajectorySearch{data/results/planning/overtaking/}{results_trajectory_1.dat}{results_problem_info.dat}{\vmin}{\vmax}
		
		\PlotMovingObjectZero{data/results/planning/overtaking/objects/results_moving_obstacle_files}{9}		
		
		\end{axis}
	\end{tikzpicture}
	\begin{tikzpicture}
		\begin{axis}[
		width=\figwidth\textwidth,
		height=\figheight\textheight,
		xmin = \xmin,
		xmax = \xmax,
		ymin = \ymin,
		ymax = \ymax,
		ytick = \empty,
		axis equal,
		colormap/YlOrBr,	
		view={0}{90},
		]
		
		\PlotSituation{data/results/planning/overtaking/}{results_obstacles.dat}{results_poly.dat}{\xmin}{\xmax}{\ymin}{\ymax}
		
		\PlotVoronoi{data/}{results/planning/overtaking/results_voronoi_path_14_filelist.dat}
		
		\PlotTrajectorySearchBranches{data/results/planning/overtaking/}{results_branch_files.dat}		
		
		\PlotTrajectorySearch{data/results/planning/overtaking/}{results_trajectory_2.dat}{results_problem_info.dat}{\vmin}{\vmax}	
		
		\PlotMovingObjectZero{data/results/planning/overtaking/objects/results_moving_obstacle_files}{15}
		
		\end{axis}
	\end{tikzpicture}	
	\shiftedsubcaption{Overtaking a slowly moving object.}	
	\label{fig:results_b}
	\end{subfigure}
	\begin{subfigure}{1.0\textwidth}
	
	\newcommand{\xmin}{0}
	\newcommand{\xmax}{25}
	\newcommand{\ymin}{-5}
	\newcommand{\ymax}{14}

	\begin{tikzpicture}
		\begin{axis}[
		width=\figwidth\textwidth,
		height=\figheight\textheight,
		xmin = \xmin,
		xmax = \xmax,
		ymin = \ymin,
		ymax = \ymax,
		axis equal,
		colormap/YlOrBr,
		view={0}{90},
		]
		
		\PlotSituation{data/results/planning/turnevasion/}{results_obstacles.dat}{results_poly.dat}{\xmin}{\xmax}{\ymin}{\ymax}
		
		\PlotVoronoi{data/}{results/planning/turnevasion/results_voronoi_path_2_filelist.dat}
		
		\PlotTrajectorySearchBranches{data/results/planning/turnevasion/}{results_branch_files.dat}		

		\PlotTrajectorySearch{data/results/planning/turnevasion/}{results_trajectory_0.dat}{results_problem_info.dat}{\vmin}{\vmax}		
		
		\PlotMovingObjectZero{data/results/planning/turnevasion/objects/results_moving_obstacle_files}{3}
		
		\PlotMovingObjectOne{data/results/planning/turnevasion/objects/results_moving_obstacle_files}{3}
		
		\PlotMovingObjectTwo{data/results/planning/turnevasion/objects/results_moving_obstacle_files}{3}		
		
		\node[draw, fill=\textboxcolor, anchor=north west] at (axis cs: \xmin,14.5) {\tiny $\SI{10}{\kilo\meter\per\hour}$};
	
		\end{axis}
	\end{tikzpicture}
	\begin{tikzpicture}

		\begin{axis}[
		width=\figwidth\textwidth,
		height=\figheight\textheight,
		xmin = \xmin,
		xmax = \xmax,
		ymin = \ymin,
		ymax = \ymax,
		ytick = \empty,
		axis equal,
		colormap/YlOrBr,
		view={0}{90},
		]
		
		\PlotSituation{data/results/planning/turnevasion/}{results_obstacles.dat}{results_poly.dat}{\xmin}{\xmax}{\ymin}{\ymax}
		
		\PlotVoronoi{data/}{results/planning/turnevasion/results_voronoi_path_7_filelist.dat}		
		\PlotTrajectorySearchBranches{data/results/planning/turnevasion/}{results_branch_files.dat}			
		
		\PlotTrajectorySearch{data/results/planning/turnevasion/}{results_trajectory_1.dat}{results_problem_info.dat}{\vmin}{\vmax}	
		
		\PlotMovingObjectZero{data/results/planning/turnevasion/objects/results_moving_obstacle_files}{8}
		
		\PlotMovingObjectOne{data/results/planning/turnevasion/objects/results_moving_obstacle_files}{8}
		
		\PlotMovingObjectTwo{data/results/planning/turnevasion/objects/results_moving_obstacle_files}{8}
		
		\end{axis}
	\end{tikzpicture}
	\begin{tikzpicture}	
	
		\begin{axis}[
		width=\figwidth\textwidth,
		height=\figheight\textheight,
		xmin = \xmin,
		xmax = \xmax,
		ymin = \ymin,
		ymax = \ymax,
		ytick = \empty,
		axis equal,
		colormap/YlOrBr,
		view={0}{90},
		]
		
		\PlotSituation{data/results/planning/turnevasion/}{results_obstacles.dat}{results_poly.dat}{\xmin}{\xmax}{\ymin}{\ymax}

		\PlotVoronoi{data/}{results/planning/turnevasion/results_voronoi_path_21_filelist.dat}		
		
		\PlotTrajectorySearchBranches{data/results/planning/turnevasion/}{results_branch_files.dat}			
		
		\PlotTrajectorySearch{data/results/planning/turnevasion/}{results_trajectory_2.dat}{results_problem_info.dat}{\vmin}{\vmax}		
		
		\PlotMovingObjectZero{data/results/planning/turnevasion/objects/results_moving_obstacle_files}{22}
		
		\PlotMovingObjectOne{data/results/planning/turnevasion/objects/results_moving_obstacle_files}{22}
		
		\PlotMovingObjectTwo{data/results/planning/turnevasion/objects/results_moving_obstacle_files}{22}

		\end{axis}
	\end{tikzpicture}
	\lastshiftedsubcaption{Turning and avoiding objects.}
	\label{fig:results_c}
	\end{subfigure}
	\end{minipage}
	\hspace{-0.4cm}
	\begin{minipage}[c]{0.09\linewidth}
	\begin{tikzpicture}
		\begin{axis}[
   	 	hide axis,
		scale only axis,
	  	point meta min=\vmin,
		point meta max=\vmax,
		colormap/YlOrBr,
		colorbar,
	 	colorbar style={
			width=0.3cm,
			height=4.5cm,
			xshift=-84mm,
			yshift=-16mm,
			ylabel=speed $v$ [\SI{}{\kilo\metre\per\hour}],
	 	 	y label style={yshift=3.0em, xshift=9.6em},
		}]
		
		\addplot[] coordinates {(0,0)};
		
		\end{axis}
	\end{tikzpicture}
	\end{minipage}
	\caption{Solutions of the time-dependent hybrid-state \astar for exemplary situations. The vehicle's speed is represented by the color gradient of the trajectory. The text boxes show the desired speed $v_\text{set}$. Gray circles: ego-vehicle. Red circles: Moving objects. Green lines: Voronoi path. Orange lines: Free-space polygon. Light blue lines: Expanded Nodes. The axis coordinates are in meters.}
	\label{fig:results}
\end{figure}
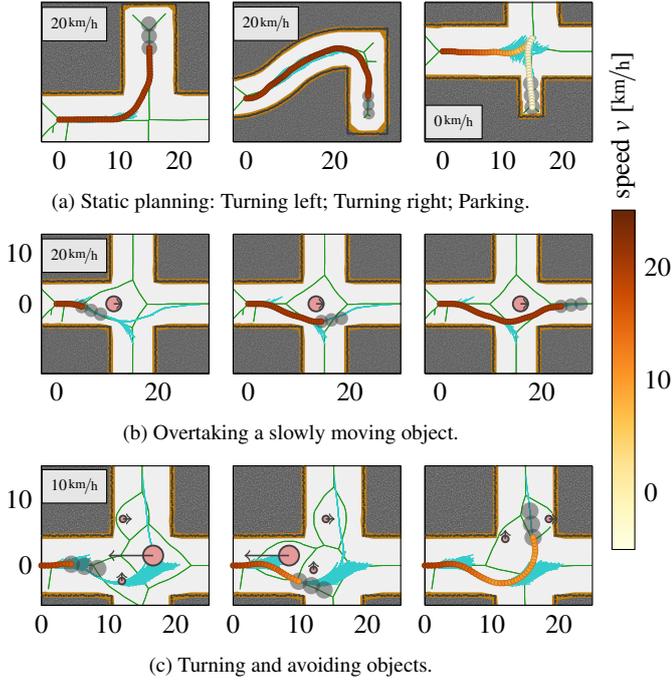

%% file: sections/results_ocp_plots.tex
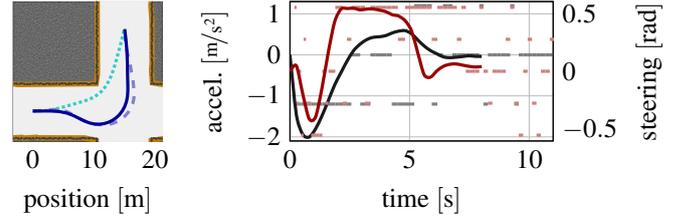
\begin{figure}[t]
	
	\pgfplotsset{xtick style={draw=none}, ytick style={draw=none}}
	
	\newcommand{\figheight}{0.14}
	\newcommand{\branchcolor}{c1}
	\newcommand{\objectcolor}{r2}
	\newcommand{\speedcolor}{k2}
	\newcommand{\vehiclecolor}{k2}
	\newcommand{\polylinewidth}{2}
	\newcommand{\voronoilinewidth}{0.3}
	\newcommand{\branchlinewidth}{0.2}
	\newcommand{\objectlinewidth}{0.7}
	\newcommand{\speedlinewidth}{0.6}
	\newcommand{\vehiclelinewidth}{0.5}
	\newcommand{\datamarksize}{0.2}
	\newcommand{\speedscale}{1.2}
	
	\newcommand{\vmin}{-5}
	\newcommand{\vmax}{25}
	
	\newcommand\noblankfileline{\expandafter\noblankaux\fileline\relax}
	\def\noblankaux#1 \relax{#1}
	
	
	\newcommand{\xmin}{-2}
	\newcommand{\xmax}{20}
	\newcommand{\ymin}{-5}
	\newcommand{\ymax}{18}
	
%
%
%
%
%
%
%
%
%
%
	\begin{tikzpicture}
	
	\begin{axis}[
	width=0.4\linewidth,
	height=\figheight\textheight,
	xmin = \xmin,
	xmax = \xmax,
	ymin = \ymin,
	ymax = \ymax,
	ytick = \empty,
	axis equal,
	colormap/YlOrBr,
	view={0}{90},
	xlabel = position,
	x unit = \si{\meter},
	]
	
	\PlotSituation{data/results/planning/turning/}{results_obstacles.dat}{results_poly.dat}{\xmin}{\xmax}{\ymin}{\ymax}
	
	\PlotRSTrajectory{data/results/planning/turning/ocp_reedsShepp_traj.dat}
	
	\PlotReferenceTraj{data/results/planning/turning/ref_moving_obs_traj.dat}
	
	\PlotTrajectory{data/results/planning/turning/ocp_moving_obs_traj.dat}	
	
	\end{axis}
	\end{tikzpicture}
	\pgfplotstableread{data/results/planning/turning/ref_moving_obs_traj.dat}{\dataRef}
	\pgfplotstableread{data/results/planning/turning/ocp_moving_obs_traj.dat}{\dataOcp}
	\hfill
	\begin{tikzpicture}
	\pgfplotsset{
		width=0.57\linewidth,
		height = \figheight\textheight,
		legend style={%
			at={(0.99,0.99)},
			anchor=north east
		},
		legend cell align=left,
		xmin=0,
		xmax=11,
		ymax=1.3,
		ymin=-2.1,
		no markers
	}
	\begin{axis}[
		grid = major,
		xlabel = {time},
		x unit = \si{\second},
		ylabel = {accel.},
		y unit = \si{\metre\per\square\second},
	]
	
	\foreach \i in {2,...,61}{
		\pgfplotstablegetelem{\i}{t}\of{data/results/planning/turning/ref_moving_obs_traj.dat}
		\edef\Tstart{\pgfplotsretval}
		\pgfplotstablegetelem{\numexpr\i+1}{t}\of{data/results/planning/turning/ref_moving_obs_traj.dat}
		\edef\Tend{\pgfplotsretval}
		\pgfplotstablegetelem{\i}{acc}\of{data/results/planning/turning/ref_moving_obs_traj.dat}
		\edef\Y{\pgfplotsretval}
		\addplot[
		k3,
		very thick
		] coordinates {
			(\Tstart, \Y)
			(\Tend, \Y)
		};
	}
	
	\addplot [k1, very thick, smooth] table [x={t}, y={acc}] {\dataOcp};
	\end{axis}
	\begin{axis}[
	yticklabel pos=right,
	ylabel = {steering},
	y unit = \si{\radian},
	xtick=\empty,
	ymax=0.6,
	ymin=-0.6,
	]
	\foreach \i in {2,...,61}{
		\pgfplotstablegetelem{\i}{t}\of{data/results/planning/turning/ref_moving_obs_traj.dat}
		\edef\Tstart{\pgfplotsretval}
		\pgfplotstablegetelem{\numexpr\i+1}{t}\of{data/results/planning/turning/ref_moving_obs_traj.dat}
		\edef\Tend{\pgfplotsretval}
		\pgfplotstablegetelem{\i}{steer}\of{data/results/planning/turning/ref_moving_obs_traj.dat}
		\edef\Y{\pgfplotsretval}
		\addplot[
		r3,
		very thick
		] coordinates {
			(\Tstart, \Y)
			(\Tend, \Y)
		};
	}
	\addplot[r1, very thick, smooth] table [x={t}, y={steer}]  {\dataOcp};
	
	\end{axis}
	
	\end{tikzpicture}

	\caption{Solution of \eqref{eq:ocp} for the scenario given in \figref{fig:results_c}. Left: Resulting optimal path in dark blue. For comparison: The dotted cyan line shows the Reeds-Shepp path and the dashed blue line shows the \astar path. Right: Control values - acceleration (black) and steering angle (red). The \astar control is represented by the respective pale dashed lines.}
	\label{fig:ocp_results}
\end{figure}

%% file: sections/Conclusion.tex
In this work, we have proposed algorithms for correlated motion planning and control of autonomous vehicles in dynamic urban environments. As a basis for both, an algorithm for the automated generation of free-space polygons in arbitrary situations has been presented. In particular, we have shown that this allows an efficient and generic representation of lane information and static obstacles. Subsequently, the time-dependent hybrid-state \astar algorithm for model-based motion planning in non-static environments has been introduced - based on free-space polygons and dynamic Voronoi paths. We have demonstrated that it can provide short-term maneuvers in a few milliseconds even for complex scenarios, which qualifies it as a basis for making tactical decisions. Furthermore, it has been illustrated how the motion planning solution can be applied as an initial guess within a trajectory optimization step for the calculation of actual vehicle controls. Here, the comparison with Reeds-Shepp paths has shown that not only a strong reduction of the computation time is achieved, but in particular also the robustness of the optimization method for difficult situations is increased. The scope of our future work is to apply and evaluate the presented concept for decision making and control with the research vehicle from \secref{subsec:vehicle} in a real urban environment. 

%% file: sections/Appendix.tex
\appendix
\section{Hyperparameters}
\begin{table}[h]
	
	\centering
	\begin{tabular}{c|c|c!{\vrule width 2pt}c|c|c}
		\multicolumn{3}{c!{\vrule width 2pt}}{\textsc{Time-Sensitive Hybrid-State \astar}} 					& \multicolumn{3}{c}{\textsc{Free-Space Polygon}} \\\hline
		\Gape[0.1cm][0cm]{\textbf{Descript.}} 			& \textbf{Var.} 		& \textbf{Value} 			& \textbf{Descript.} 			& \textbf{Var.} 		& \textbf{Value} \\\hline
		\multirow{3}{*}{Weights} 	& $w_v$ 			& 1 												& Refinements & $\refinements$				& 2 \\
									& $w_o$ 			& 2 												& Clearance & $\clearance$					& \SI{10}{\meter} \\
									& $w_p$ 			& 1 												& Expansion & $\expansion$					& \SI{20}{\meter} \\\cline{1-3}
		\multirow{2}{*}{Voronoi} 	& $\alpha$ 			& 1000												& \multicolumn{3}{c}{} \\
				 					& $\dist_{\max}$	& \SI{4}{\meter} 									& \multicolumn{3}{c}{\textsc{Optimal Control}} \\\cline{1-6}
		\multirow{3}{*}{Controls}	& $\A$ 				& $\{0, \pm1.2\}$ \si{\meter\squared\per\second} 	&\Gape[0.1cm][0cm]{\textbf{Descript.}} 			& \textbf{Var.} 		& \textbf{Value} \\\cline{4-6}
									& \multirow{2}{*}{$\B$} & $\{0, \pm0.275, $  							& \multirow{4}{*}{Weights}								& $w_0$					& 0.3\\
									& 					& $\pm0.55\}$ \si{\radian} 							&								& $w_1$					& 0.3 \\\cline{1-3}
		\multirow{4}{*}{Discretiz.} & spatial			& \SI{0.5}{\meter} 									& 								& $w_2$					& 0.1 \\
									& yaw 				& \SI{0.1}{\radian}									&								& $w_3$					& 2.5\\\cline{4-6}
									& speed 			& \SI{0.5}{\meter\per\second} 						& \# Discrete 					& \multirow{2}{*}{--}	& \multirow{2}{*}{31}\\
									& time 				& \SI{0.3}{\second} 								& Points						& 						& \\
	\end{tabular}
\end{table}